\documentclass[sigconf]{acmart}

\renewcommand\footnotetextcopyrightpermission[1]{} 
\usepackage{colortbl}
\usepackage[normalem]{ulem}
\useunder{\uline}{\ul}{}
\usepackage[ruled]{algorithm2e}
\usepackage{subcaption}
\definecolor{ylp_color1}{RGB}{255,193,193}
\definecolor{ylp_color2}{RGB}{255,228,225}
\usepackage{tcolorbox}
\newtcbox{\mybox}[1][red]{on line, colback = {RGB}{255,228,225}, colframe = {RGB}{255,193,193},  arc=1mm, auto outer arc, boxrule=0.5pt,}
\usepackage{multirow}
\usepackage{booktabs}
\usepackage{wrapfig}
\usepackage{amsfonts}
\usepackage{bm}
\usepackage{eucal}
\usepackage{arydshln}

\newtheorem{theorem}{Theorem}
\newtheorem{lemma}{Lemma}
\newtheorem{assumption}{Assumption}
\usepackage{pifont}
\usepackage{setspace}
\usepackage{url}
\usepackage{enumitem}
\newcommand{\circled}[1]{\textcircled{\small #1}}

\newcommand{\hetero}{heterogeneous }
\newcommand{\homo}{homogeneous }
\newcommand{\pers}{personalized }
\newcommand{\gen}{generalized }

\newcommand{\heteroN}{heterogeneity }

\newcommand{\persN}{personalization }
\newcommand{\genN}{generalization }
\newcommand{\fe}{feature extractor }

\newcommand{\rep}{representation }
\newcommand{\reps}{representations }
\newcommand{\sota}{state-of-the-art }
\newcommand{\assum}{Assumption }

\newcommand{\methodname}{{\tt{pFedMoE}}}

\AtBeginDocument{%
  \providecommand\BibTeX{{%
    \normalfont B\kern-0.5em{\scshape i\kern-0.25em b}\kern-0.8em\TeX}}}

\setcopyright{acmlicensed}
\copyrightyear{2018}
\acmYear{2018}
\acmDOI{XXXXXXX.XXXXXXX}

\acmConference[Conference acronym 'XX]{Make sure to enter the correct
  conference title from your rights confirmation emai}{June 03--05,
  2018}{Woodstock, NY}
\acmBooktitle{Woodstock '18: ACM Symposium on Neural Gaze Detection,
 June 03--05, 2018, Woodstock, NY} 
\acmISBN{978-1-4503-XXXX-X/18/06}

\acmSubmissionID{46}

\begin{document}

\title[pFedMoE for Model-Heterogeneous Personalized Federated Learning]{pFedMoE: Data-Level Personalization with Mixture of Experts for Model-Heterogeneous Personalized Federated Learning}

\author{Liping Yi}
\email{yiliping@nbjl.nankai.edu.cn}
\orcid{0000-0001-6236-3673}
\affiliation{%
  \institution{College of C.S., TMCC, SysNet, DISSec, GTIISC, Nankai University}
  \city{Tianjin}
  \country{China}
}

\author{Han Yu}
\email{han.yu@ntu.edu.sg}
\orcid{0000-0001-6893-8650}
\affiliation{%
  \institution{School of Computer Science and Engineering, Nanyang Technological University (NTU)}
  \country{Singapore}
}

\author{Chao Ren}
\email{chao.ren@ntu.edu.sg}
\orcid{0000-0001-9096-8792} 
\affiliation{%
  \institution{School of Computer Science and Engineering, Nanyang Technological University (NTU)}
  \country{Singapore}
}

\author{Heng Zhang}
\email{hengzhang@tju.edu.cn}
\orcid{0000-0003-4874-6162}
\affiliation{%
  \institution{College of Intelligence and Computing, Tianjin University}
  \city{Tianjin}
  \country{China}
}

\author{Gang Wang}
\email{wgzwp@nbjl.nankai.edu.cn}
\orcid{0000-0003-0387-2501}
\affiliation{%
  \institution{College of C.S., TMCC, SysNet, DISSec, GTIISC, Nankai University}
  \city{Tianjin}
  \country{China}
}

\author{Xiaoguang Liu}
\email{liuxg@nbjl.nankai.edu.cn}
\orcid{0000-0002-9010-3278}
\affiliation{%
  \institution{College of C.S., TMCC, SysNet, DISSec, GTIISC, Nankai University}
  \city{Tianjin}
  \country{China}
}

\author{Xiaoxiao Li}
\email{xiaoxiao.li@ece.ubc.ca}
\affiliation{%
  \institution{Electrical and Computer Engineering Department, University of British Columbia (UBC)}
  \city{Vancouver}
  \country{Canada}
}

\renewcommand{\shortauthors}{L. Yi and H. Yu, et al.}

\begin{abstract}

Federated learning (FL) has been widely adopted for collaborative training on decentralized data. However, it faces the challenges of data, system, and model heterogeneity. This has inspired the emergence of model-heterogeneous personalized federated learning (MHPFL). 
Nevertheless, the problem of ensuring data and model privacy, while achieving good model performance and keeping communication and computation costs low remains open in MHPFL.
To address this problem, we propose a model-heterogeneous \underline{p}ersonalized \underline{Fed}erated learning with \underline{M}ixture \underline{o}f \underline{E}xperts (\methodname{}) method. It assigns a shared homogeneous small feature extractor and a local gating network for each client's local heterogeneous large model.
Firstly, during local training, the local heterogeneous model's feature extractor acts as a \textit{local expert} for personalized feature (representation) extraction, while the shared homogeneous small feature extractor serves as a \textit{global expert} for generalized feature extraction. The \textit{local gating network} produces \pers weights for extracted \reps from both experts on each data sample. The three models form a local \hetero \textit{MoE}. The weighted mixed \rep fuses generalized and personalized features and is processed by the local \hetero large model's header with \pers prediction information. The MoE and prediction header are updated simultaneously.
Secondly, the trained local homogeneous small feature extractors are sent to the server for cross-client information fusion via aggregation. 
Overall, \methodname{} enhances local model personalization at a fine-grained data level, while supporting model heterogeneity.
We theoretically prove its convergence over time. Extensive experiments over $2$ benchmark datasets and $7$ existing methods demonstrate its superiority with up to $2.80\%$ and $22.16\%$ accuracy improvement over the state-of-the-art and the same-category best baselines, while incurring lower computation and satisfactory communication costs. 
\end{abstract}

\maketitle

\section{Introduction}

Federated learning (FL) \cite{FedAvg,1w-survey} is a distributed machine learning paradigm supporting collaborative model building in a privacy-preserving manner. In a typical FL algorithm - {\tt{FedAvg}} \cite{FedAvg}, an FL server selects a subset of FL clients (i.e., data owners), and sends them the global model. Each selected client initializes its local model with the received global model, and trains it on its local data. The trained local models are then uploaded to the server for aggregation to generate a new global model by weighted averaging. Throughout this process, only model parameters are exchanged between the server and clients, thereby avoiding exposure to potentially sensitive local data. This paradigm requires clients and the server to maintain the same model structure (i.e., model homogeneity).

In practice, FL faces challenges related to various types of heterogeneity. Firstly, decentralized data from clients are often non-independent and identically distributed (non-IID), \emph{i.e.}, \textbf{data or statistical heterogeneity}. A single shared global model trained on non-IID data might not adapt well to each client's local data distribution.
Secondly, in cross-device FL, clients are often mobile edge devices with diverse system configurations (\emph{e.g.}, bandwidth, computing power), \emph{i.e.}, \textbf{system heterogeneity}. If all clients share the same model structure, the model size must be compatible with the lowest-end device, causing performance bottlenecks and resource wastage on high-end devices.
Thirdly, in cross-silo FL, clients are institutions or enterprises concerned with protecting model intellectual property and maintaining different private model repositories, \emph{i.e.}, \textbf{model heterogeneity}. Their goal is often to further train existing proprietary models through FL without revealing them. 
Therefore, the field of Model-Heterogeneous Personalized Federated learning (MHPFL) has emerged, aiming to train personalized and heterogeneous local models for each FL client.

Existing MHPFL methods supporting completely \hetero models can be divided into three categories: (1) knowledge distillation-based MHPFL \citep{FedProto}, which either relies on extra public data with similar distributions as local data
or incurs additional computational and communication burdens on clients to perform knowledge distillation; (2) model mixup-based MHPFL \citep{LG-FedAvg}, which splits client models into shared homogeneous and private heterogeneous parts, but sharing only the homogeneous part bottlenecks model performances, revealing model structures in the process; and (3) mutual learning-based MHPFL \citep{FedKD}, which alternately trains private heterogeneous large models and shared homogeneous small models for each client in a mutual learning manner, incurring additional computation costs for clients. 

\begin{figure}[!t]
\centering
\includegraphics[width=0.6\linewidth]{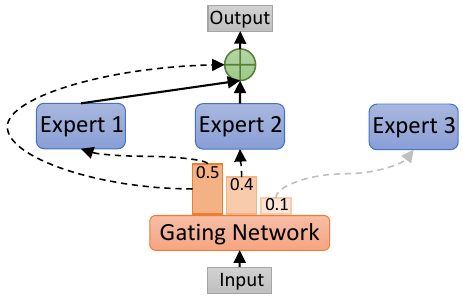}
\caption{Workflow of MoE.}
\vspace{-1em}
\label{fig:MoE}
\vspace{-1em}
\end{figure}

With the rapid development of large language models (LLMs), incorporating multiple data modalities like images and text to train such models increases training and inference costs. Besides increasing LLMs scales or fine-tuning, the Mixture of Experts (MoE) approach has shown promise to address this issue. An MoE (Figure~\ref{fig:MoE}) consists of a gating network and multiple expert models. During training, a data sample passes through the gating network to produce weights for all experts. The top-$p$ weighted experts process this sample. Their predictions, weighted by their corresponding weights, form the final output. The loss between the mixed output and the label is used to update the $p$ experts and the gating network simultaneously. The key idea of MoE is to partition data into subtasks using the gating network and assign specific experts to handle different subtasks based on their expertise. This allows MoE to address both general and specialized problems. 
Existing FL MoE methods only address data \heteroN in typical model-\homo FL settings by allowing a client to use the gating network either for selecting specific local models from other clients or for balancing the global and local models.

Previous study \cite{FedCP} highlights that each data sample contains both generalized and personalized information, with proportions varying across samples. Inspired by this insight, we propose the model-heterogeneous \underline{p}ersonalized \underline{Fed}erated learning with \underline{M}ixture \underline{o}f \underline{E}xperts (\methodname{}) method, to enhance personalization at the data level to address data \heteroN and support model heterogeneity.
Under \methodname{}, each FL client's model consists of a local gating network, a local heterogeneous large model's feature extractor (\emph{i.e.}, the local expert) for personalized information extraction, and a globally shareable homogeneous small feature extractor (\emph{i.e.}, the global expert) for extracting generalized information, thereby forming a local MoE. During local training, for each local data sample, the gating network adaptively produces personalized weights for the representations extracted by the two experts. The weighted mixed representation,  incorporating both \gen and \pers feature information, is then processed by the local heterogeneous model's prediction header infused with personalized predictions. The hard loss between predictions and labels simultaneously updates MoE and the header. 
After local training, the homogeneous small feature extractors are sent to the FL server to facilitate knowledge sharing among heterogeneous local models.

Theoretical analysis proves that \methodname{} can converge over time.
Extensive experiments on $2$ benchmark datasets and $7$ existing methods demonstrate that \methodname{} achieves state-of-the-art model accuracy, while incurring lower computational and acceptable communication costs. Specifically, it achieves up to $2.80\%$ and $22.16\%$ higher test accuracy over the state-of-the-art and the same-category best baselines, respectively.



\section{Related Work}\label{sec:related}
\subsection{Model-Heterogeneous Personalized FL}
Existing MHPFL has two families: (1) Clients train \textit{heterogeneous local subnets} of the global model by model pruning, and the server aggregates them by parameter ordinate, \emph{e.g.}, {\tt{FedRolex}} \citep{FedRolex}, {\tt{FLASH}}~\citep{FLASH}, {\tt{HeteroFL}} \citep{HeteroFL}, {\tt{FjORD}} \citep{FjORD}, {\tt{HFL}} \citep{HFL}, {\tt{Fed2}} \citep{Fed2}, {\tt{FedResCuE}} \citep{FedResCuE}; (2) Clients hold \textit{completely heterogeneous local models} and exchange knowledge with others by knowledge distillation, model mixture, and mutual learning. We focus on the second category, which supports high model heterogeneity and is more common in practice.

\textbf{MHPFL with Knowledge Distillation.} 
Some methods utilize knowledge distillation on an additional (labeled or unlabeled) public dataset with a similar distribution as local data at the server or clients to fuse across-client information, such as 
{\tt{Cronus}} \citep{Cronus}, {\tt{FedGEMS}} \citep{FedGEMS}, {\tt{Fed-ET}} \citep{Fed-ET}, {\tt{FSFL}} \citep{FSFL}, {\tt{FCCL}} \citep{FCCL}, {\tt{DS-FL}} \citep{DS-FL}, {\tt{FedMD}} \citep{FedMD}, {\tt{FedKT}} \citep{FedKT}, {\tt{FedDF}} \citep{FedDF}, {\tt{FedHeNN}} \citep{FedHeNN}, {\tt{FedKEM}} \citep{FedKEM}, {\tt{KRR-KD}} \citep{KRR-KD}, {\tt{FedAUX}} \citep{FEDAUX}, {\tt{CFD}} \citep{CFD}, {\tt{pFedHR}} \citep{pFedHR}, {\tt{FedKEMF}} \citep{FedKEMF} and {\tt{KT-pFL}} \citep{KT-pFL}) 
However, obtaining such a public dataset is difficult due to data privacy. Distillation on clients burdens computation, while communicating logits or representation of each public data sample between the server and clients burdens communication. 
To avoid using public data, {\tt{FedGD}} \citep{FedGD}, {\tt{FedZKT}} \citep{FedZKT} and {\tt{FedGen}} \citep{FedGen} train a global generator to produce synthetic data for replacing public data, but generator training is time-consuming and reduces FL efficiency. {\tt{HFD}} \citep{HFD1,HFD2}, {\tt{FedGKT}} \citep{FedGKT}, {\tt{FD}} \citep{FD}, {\tt{FedProto}} \citep{FedProto}, and {\tt{FedGH}} \citep{FedGH} do not rely on public or synthetic data. Instead, clients share seen classes and corresponding class-average logits or representations with the server, which are then distilled with global logits or representations of each class. However, they incur high computation costs on clients, and might be restricted in privacy-sensitive scenarios due to class uploading.

\textbf{MHPFL with Model Mixture.} 
A local model is split into a feature extractor and a classifier. {\tt{FedMatch}} \citep{FedMatch}, {\tt{FedRep}} \citep{FedRep}, {\tt{FedBABU}} \citep{FedBABU} and {\tt{FedAlt/FedSim}} \citep{FedAlt/FedSim} share \homo feature extractors, while holding \hetero classifiers. {\tt{FedClassAvg}} \citep{FedClassAvg}, {\tt{LG-FedAvg}} \citep{LG-FedAvg} and {\tt{CHFL}} \citep{CHFL} behave oppositely. They inherently only offer models with partial heterogeneity, potentially leading to performance bottlenecks and partial model structure exposure.

\textbf{MHPFL with Mutual Learning.} 
Each client in {\tt{FML}} \citep{FML} and {\tt{FedKD}} \citep{FedKD} has a local \hetero large model and a shareable \homo small model, which are trained alternately via mutual learning. The trained \homo small models are aggregated at the server to fuse information from different clients. However, alternative training increases computational burdens. Recent {\tt{FedAPEN}} \citep{FedAPEN} improves {\tt{FML}} by enabling each client to first learn a trainable weight $\lambda$ for local \hetero model outputs, with ($1-\lambda$) is assigned to the shared \homo model outputs; then fixing this pair of weights and training two models with the ensemble loss between the weighted ensemble outputs and labels. Due to diverse data distributions among clients, the learnable weights are diverse, \emph{i.e.}, achieving \textbf{client-level personalization}. Whereas, it fails to explore both \gen and \pers knowledge at the data level due to fixing weights during training.

\textbf{Insight}.
In contrast, our proposed \methodname{} treats the shareable \homo small feature extractor and the local \hetero large model's feature extractor as global and local experts of an MoE. It deploys a lightweight linear gating network to produce \pers weights for the representations of both experts for each data sample, enabling the extraction of both global \gen and local \pers knowledge at a more fine-grained \textbf{data-level personalization} that adapts to in-time data distribution. 
Besides, \methodname{} simultaneously updates three models in MoE, saving training time compared to first training the learnable weights and then alternately training models as in {\tt{FedAPEN}}. 
Clients and the server in \methodname{} only exchange \homo small feature extractors, thereby reducing communication costs and preserving local data and model privacy.

\subsection{MoE in Federated Learning}
To address the data heterogeneity issue in typical FL,
{\tt{FedMix}}~\citep{FedMix} and {\tt{FedJETs}}~\citep{FedJETs} allow each client to construct an MoE with a shared gating network and a homogeneous local model. The gating network selects specific other local models more adaptive to this client's local data for ensembling. These methods incur significant communication costs as they send the entire model to each client.
There is also a PFL method~\cite{xxl-moe} using MoE for domain adaption across non-IID datasets.
\citet{FL-MoE2} and {\tt{PFL-MoE}} \citep{PFL-MoE} incorporated MoE into \pers FL to mitigate data heterogeneity in model-\homo scenarios. In each round, each client receives the global model from the server as a global expert and fine-tunes it on partial local data as a local expert, the two experts and a gating network form a MoE. During MoE training, each client utilizes a \pers gating network with only one linear layer to produce weights of the outputs of two experts. Then the weighted output is used for updating the local model and the gating network on the remaining local data. 
Although alleviating data heterogeneity through data-level personalization, they face two constraints: (1) training MoE on partial local data may compromise model performances, and (2) the one-linear-layer gating network with fewer parameters extracts only limited knowledge from local data.

In contrast, \methodname{} enhances data-level personalization in the more challenging model-heterogeneous FL scenarios. The gating network in \methodname{} produces weights for the two experts' representations, thereby carrying more information than outputs and facilitating the fusion of global \gen and local \pers features. The weighted mixed representations are processed by the prediction header of the local \pers \hetero models to enhance prediction personalization. We devise a more efficient gating network to learn local data distributions. We train the three models of MoE simultaneously on all local data, boosting model performances and saving training time. Only the small shared \homo feature extractors are transmitted, thereby incurring low communication costs.

\section{Preliminaries}
Consider a typical model-\homo FL algorithm (e.g., {\tt{FedAvg}} \citep{FedAvg}) for an FL system comprising a server and $N$ clients. In each communication round: 1) the server selects $K=C\cdot N$ clients $\boldsymbol{\mathcal{S}}$ ($C$ is sampling ratio, $K$ is the number of selected clients, $\boldsymbol{\mathcal{S}}$ is the selected client set, and $|\boldsymbol{\mathcal{S}}|=K$) and broadcasts the global model $\mathcal{F}(\omega)$ (where $\mathcal{F}$ is the model structure, and $\omega$ are the model parameters) to the selected clients. 2) A client $k$ initializes its local model $\mathcal{F}(\omega_k)$ with the received global model $\mathcal{F}(\omega)$, and trains it on its local data $D_k$ (where $D_k \sim P_k$ indicates that the data from different clients follow non-IID distributions) by $\omega_k\gets\omega_k-\eta_\omega\nabla\ell(\mathcal{F}({\boldsymbol{x}_i;\omega}_k),y_i)$, $(\boldsymbol{x}_i,y_i)\in D_k$. Then, the updated local model ${\mathcal{F}(\omega}_k)$ is uploaded to the server. 3) The server aggregates the received local models to produce a global model by $\omega=\sum_{k\in\boldsymbol{\mathcal{S}}}\frac{n_k}{n}\omega_k$ ($n_k=|D_k|$, the sample size of client-$k$'s local data $D_k$; $n$ is sample size across all clients). The above steps are repeated until the global model converges. Typical FL aims to minimize the average loss of the global model on local data across all clients:
\begin{equation}
\min _\omega \sum_{k=0}^{N-1} \frac{n_k}{n} \ell(\mathcal{F}(D_k ; \omega)).
\end{equation}
This definition requires that all clients and the server must possess models with identical structures $\mathcal{F}(\cdot )$, \emph{i.e.}, \textbf{model-homogeneous}.

\methodname{} is designed for model-heterogeneous PFL for supervised learning tasks. We define client $k$'s local \hetero model as $\mathcal{F}_k(\omega_k)$ ($\mathcal{F}_k(\cdot)$ is the heterogeneous model structure; $\omega_k$ are the \pers model parameters). The objective is to minimize the sum of the loss of local \hetero models on local data:
\begin{equation}\label{eq:FedMoE1}
\min _{\omega_{0, \ldots, N-1}} \sum_{k=0}^{N-1} \ell(\mathcal{F}_k(D_k ; \omega_k)).
\end{equation}

\section{The Proposed Approach}
\textbf{Motivation.} In FL, the global model has ample \gen knowledge, while local models have \pers knowledge. Participating clients, with limited local data, hope to enhance the generalization of their local models to improve model performances. For a client $k$, its local \hetero model $\mathcal{F}_k(\omega_k)$ comprises a feature extractor $\mathcal{F}_k^{ex}(\omega_k^{ex})$ and a prediction header $\mathcal{F}_k^{hd}(\omega_k^{hd})$, $\mathcal{F}_k(\omega_k)=\mathcal{F}_k^{ex}(\omega_k^{ex})\circ\mathcal{F}_k^{hd}(\omega_k^{hd})$. The feature extractor captures low-level \pers feature information, while the prediction header incorporates high-level \pers prediction information. 
Hence, \textbf{(1) we enhance the generalization of the local heterogeneous feature extractor} to extract more generalized features through FL, while retaining the prediction header of the local \hetero model to enhance \pers prediction capabilities.
Furthermore, \citet{FedCP} highlighted that various local data samples of a client contain differing proportions of global \gen information and local \pers information. This motivates us to \textbf{(2) dynamically balance the \genN and \persN of local \hetero models, adapting to non-IID data across different clients at the data level}.

\begin{figure}[t]
\centering
\includegraphics[width=\linewidth]{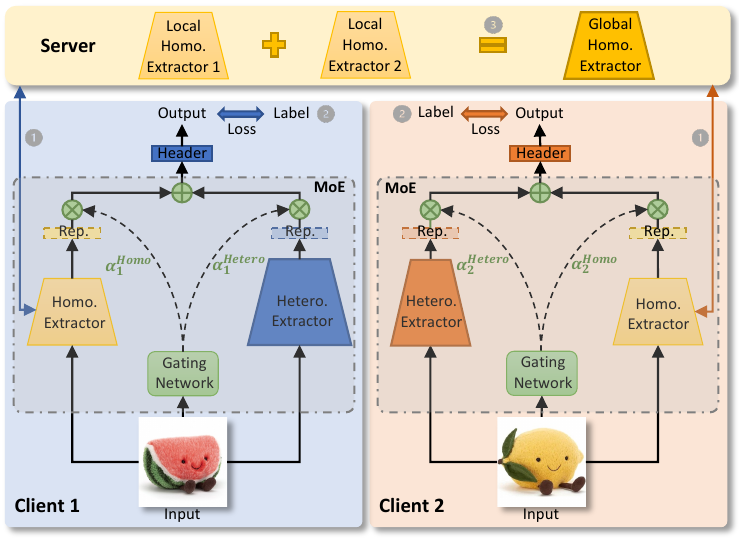}
\vspace{-1em}
\caption{Workflow of \methodname{}.}
\label{fig:FedMoE}
\vspace{-1em}
\end{figure}

\textbf{Overview.} To realize the above insights, \methodname{} incorporates a shareable small \homo feature extractor $\mathcal{G}(\theta)$ far smaller than the local \hetero feature extractor $\mathcal{F}_k^{ex}(\omega_k^{ex})$. As shown in Figure.~\ref{fig:FedMoE}, in the $t$-th communication round, the workflow of \methodname{} involves the following steps:
\begin{enumerate}[label=\circled{\arabic*}]
    \item The server samples $K$ clients $\boldsymbol{\mathcal{S}^t}$ and sends the global \homo small \fe $\mathcal{G}(\theta^{{t}-{1}})$ aggregated in the $(t-1)$-th round to them.
    \item Client $k \in \boldsymbol{\mathcal{S}^t}$ regards the received global \homo small \fe $\mathcal{G}(\theta^{{t}-{1}})$ as the global expert for extracting \gen feature across all classes, and treats the local \hetero large \fe $\mathcal{F}_k^{ex}(\omega_k^{ex,t-1})$ as the local expert for extracting \pers feature of local seen classes. A \homo or heterogeneous lightweight \pers local gating network $\mathcal{H}(\varphi_k^{t-1})$ is introduced to balance \genN and \persN by dynamically producing weights for each sample's representations from two experts. The three models form an MoE architecture. The weighted mixed representation from MoE is then processed by the local \hetero large model's prediction header $\mathcal{F}_k^{hd}(\omega_k^{hd,t-1})$ to extract \pers prediction information. The three models in MoE and the header are trained simultaneously in an end-to-end manner. The updated \homo $\mathcal{G}(\theta_k^t)$ is uploaded to the server, while $\mathcal{F}_k(\omega_k^t)$, $\mathcal{H}(\varphi_k^t)$ are retained by the clients.
    \item The server aggregates the received local \homo feature extractors $\mathcal{G}(\theta_k^t)$ ($k \in \boldsymbol{\mathcal{S}^t}$) by weighted averaging to produce a new global \homo \fe $\mathcal{G}(\theta^{t})$.
\end{enumerate}
The above process iterates until all local \hetero complete models (MoE and prediction header) converge. At the end of FL, local \hetero complete models are used for inference. The details of \methodname{} are illustrated in Algorithm~\ref{alg:FedMoE} (Appendix~\ref{app:pseudo-codes}).

\subsection{MoE Training}
In the MoE, each local data sample $(\boldsymbol{x}_i,y_i)\in D_k$ is fed into the global expert $\mathcal{G}(\theta^{{t}-\mathbf{1}})$ to produce the \textit{\gen representation}, and simultaneously into the local expert $\mathcal{F}_k^{ex}(\omega_k^{ex,t-1})$ to generate the \textit{\pers representation},
\begin{equation}
\boldsymbol{\mathcal{R}}_{k, i}^{\mathcal{G}, t}=\mathcal{G}(\boldsymbol{x}_i ; \theta^{t-1}), 
\boldsymbol{\mathcal{R}}_{k, i}^{\mathcal{F}_{k},t}=\mathcal{F}_k^{e x}(\boldsymbol{x}_i ; \omega_k^{e x, t-1}).
\end{equation}
Each local data sample $(\boldsymbol{x}_i,y_i)\in D_k$ is also fed into the local gating network $\mathcal{H}(\varphi_k^{t-1})$ to produce weights for the two experts,
\begin{equation}\label{eq:gating}
[\alpha_{k, i}^{\mathcal{G}, t}, \alpha_{k, i}^{\mathcal{F}_{k}, t}]=\mathcal{H}(\boldsymbol{x}_i ; \varphi_k^{t-1}), s.t. \ 
\alpha_{k, i}^{\mathcal{G}, t}+\alpha_{k, i}^{\mathcal{F}_{k}, t}=1.
\end{equation}
Notice that different clients can hold \hetero gating networks $\mathcal{H}_k(\varphi_k)$, with the same input dimension $d$ as the local data sample $\boldsymbol{x}$ and the same output dimension $h=2$. For simplicity of discussion, we use the same gating network $\mathcal{H}(\varphi_k)$ for all clients.

Then, we mix the \reps of two experts with the weights produced by the gating network,
\begin{equation}
\boldsymbol{\mathcal{R}}_{k, i}^t=\alpha_{k, i}^{\mathcal{G}, t} \cdot \boldsymbol{\mathcal{R}}_{k, i}^{\mathcal{G}, t}+\alpha_{k, i}^{\mathcal{F}_{k}, t} \cdot \boldsymbol{\mathcal{R}}_{k, i}^{\mathcal{F}_{k},t}.    
\end{equation}
To enable the above \rep mixture, we require that the last layer dimensions of the \homo small feature extractor and the \hetero large feature extractor are identical.
The mixed \rep $\boldsymbol{\mathcal{R}}_{k,i}^t$ is then processed by the local \pers perdition header $\mathcal{F}_k^{hd}(\omega_k^{hd,t-1})$ (both \homo and \hetero headers are allowed, we use \homo headers in this work) to produce the prediction,
\begin{equation}
\hat{y}_i=\mathcal{F}_k^{h d}(\boldsymbol{\mathcal{R}}_{k, i}^t ; \omega_k^{h d, t-1}).
\end{equation}

We compute the hard loss (\emph{e.g.} Cross-Entropy loss \citep{CEloss}) between the prediction and the label as:
\begin{equation}
 \ell_i=C E(\hat{y}_i, y_i).   
\end{equation}
Then, we update all models simultaneously via gradient descent (\emph{e.g.}, SGD optimizer \citep{SGD}) in an end-to-end manner,
\begin{equation}
\begin{aligned}
\theta_k^t &\gets \theta^{t-1}-\eta_\theta \nabla \ell_i, \\
\omega_k^t &\gets \omega_k^{t-1}-\eta_\omega \nabla \ell_i, \\
\varphi_k^t &\gets \varphi_k^{t-1}-\eta_{\varphi} \nabla \ell_i,    
\end{aligned}
\end{equation}
where $\eta_\theta,\eta_\omega, \eta_\varphi$ are the learning rates of the \homo small feature extractor, the \hetero  large model, and the gating network. To enable stable convergence, we set $\eta_\theta = \eta_\omega$.

\subsection{Homogeneous Extractor Aggregation}
After local training, $k$ uploads its local \homo small feature extractor $\theta_k^t$ to the server. The server then aggregates them by weighted averaging to produce a new global feature extractor:
\begin{equation}
\theta^t=\sum_{k \in \boldsymbol{\mathcal{S}^t}} \frac{n_k}{n} \theta_k^t. 
\end{equation}

\textbf{Problem Re-formulation.} The local \pers gating networks of different clients dynamically produce weights for the \reps of two experts on each sample of local non-IID data, balancing \genN and \persN based on local data distributions. Thus, \methodname{} enhances \persN of model-\hetero \pers FL at the fine-grained \textbf{data level}. Therefore, the objective defined in Eq.~(\ref{eq:FedMoE1}) can be specified as:
\begin{equation}
\footnotesize
\min_{\omega_{0, \ldots, N-1}} \sum_{k=0}^{N-1} \ell(\mathcal{F}_k^{h d}((\mathcal{G}(D_k ; \theta), \mathcal{F}_k^{ex}(D_k ; \omega_k^{e x})) \cdot \boldsymbol{\mathcal{H}({D}_k ; {\varphi}_k)} ; \omega_k^{h d})).  
\end{equation}
$\boldsymbol{\mathcal{H}({D}_k ; {\varphi}_k)}$ denotes the weights $[\alpha^\mathcal{G},\alpha^{\mathcal{F}_k}]$ of two experts. $\cdot$ is dot product (\emph{i.e.}, summing after element-wise multiplication).

\subsection{Gating Network Design}
\begin{figure}[!t]
\centering
\includegraphics[width=1\linewidth]{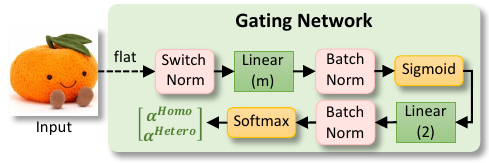}
\vspace{-2em}
\caption{Gating network structure.}
\label{fig:GatingNetwork}
\vspace{-2em}
\end{figure}

The local gating network $\mathcal{H}(\varphi_k)$ takes each data sample $\boldsymbol{x}_i\in D_k$ as the input, and outputs two weights ${[\alpha}_{k,i}^{\mathcal{G}},\alpha_{k,i}^{\mathcal{F}_k}]$ (summing to $1$) for the \reps of the two experts, as defined in Eq.~\eqref{eq:gating}. A linear network is the simplest model to fulfill these functions. Therefore, we customize a dedicated lightweight linear gating network for \methodname{}, depicted in Figure~\ref{fig:GatingNetwork}.

\textbf{Linear Layer.} \methodname{} trains models in batches. When processing a batch of color image samples, the input dimension is $[length, width,$ $channel=3, batch size]$. Before feeding it into the gating network, we flatten it to a vector with $[(length \cdot width \cdot 3, batch size])$ pixels. Given the large input vector, a gating network with only one linear layer containing $2$ neurons might not efficiently capture local data knowledge and could be prone to overfitting due to limited parameter capacity. Hence, we employ $2$ linear layers for the gating network: the first layer with $m$ neurons ($length \cdot width \cdot 3 \cdot m$ parameters), and the second layer with $2$ neurons ($m \cdot 2$ parameters).

\textbf{Normalization.} Normalization techniques are commonly employed in deep neuron networks for regularization to improve model generalization and accelerate training. Common approaches include batch, instance, and layer normalization. Recently, switch normalization \cite{switchnorm} integrates the advantages of these typical methods and efficiently handles batch data with diverse characteristics \cite{pFedGate}. After flattening the input, we apply a switch normalization layer before feeding it into the first linear layer. To leverage the benefits of widely adopted batch normalization, we include batch normalization layers after two linear layers.

\textbf{Activation Function.} Activation functions increase non-linearity to improve deep network expression, mitigating gradient vanishing or explosion. Commonly used activation functions include Sigmoid, ReLU, and Softmax, each with its range of values. Since the gating network's output weights range between $0$ and $1$, we add a Sigmoid activation layer after the first linear layer to confine its output within $(0,1)$. We add a Softmax activation layer after the second linear layer to ensure that the produced two weights sum to $1$.

\subsection{Discussion}

Here, we further discuss the following aspects of \methodname{}.

\indent \textbf{Privacy.} Clients share the \homo small feature extractors for knowledge exchange. Local \hetero large models and local data remain with the clients, thereby preserving their privacy.

\textbf{Communication.} Only \homo small feature extractors are transmitted between the server and clients, incurring lower communication costs than transmitting complete models as in {\tt{FedAvg}}.

\textbf{Computation.} Apart from training local \hetero large models, clients also train a small \homo feature extractor and a lightweight linear gating network. However, due to their smaller sizes than the \hetero large feature extractor, the computation costs are acceptable. Moreover, simultaneous training of MoE and the prediction header reduces training time.

\section{Analysis}
We first clarify additional notations used for analysis in Table~\ref{tab:notion}.


\begin{table}[!t]
\vspace{-1em}
\centering
\caption{Description of Additional Notations.}
\vspace{-1em}
\label{tab:notion}
\resizebox{\linewidth}{!}{%
\begin{tabular}{|c|p{0.95\linewidth}|}
\hline
\textbf{Notation}        & \multicolumn{1}{c|}{\textbf{Description}}                                                                                                                                                                     \\ \hline
$t \in \{0,\ldots,T-1\}$ & communication round                                                                                                                                                                                           \\ \hline
$e\in\{0,1,\ldots,E\}$   & local iteration                                                                                                                                                                                               \\ \hline
$tE+0$                   & before the $(t+1)$-th round's local training, client $k$ receives the global \homo small feature extractor $\mathcal{G}(\theta^{t})$ aggregated in the $t$-th round                                           \\ \hline
$tE+e$                   & the $e$-th local iteration in the $(t+1)$-th round                                                                                                                                                            \\ \hline
$tE+E$                   & the last local iteration, after that, client $k$ uploads the local \homo small feature extractor to the server                                                                                                \\ \hline
$W_k$                    & client $k$'s local complete model involving the MoE $\{\mathcal{G}(\theta), \mathcal{F}_k^{ex}(\omega_k^{ex}),\mathcal{H}(\varphi_k)\}$ and the perdition header $\mathcal{F}_k^{hd}(\omega_k^{hd})$ \\ \hline
$\eta$                   & the learning rate of the client $k$'s local complete model $W_k$, involving $\{\eta_\theta,\eta_\omega,\eta_\varphi\}$                                                                                        \\ \hline
\end{tabular}%
}
\vspace{-1em}
\end{table}

\begin{assumption}\label{assump:Lipschitz}
\textbf{Lipschitz Smoothness}. Gradients of client $k$'s local complete heterogeneous model $W_k$ are $L1$--Lipschitz smooth \cite{FedProto},
\begin{equation}\label{eq:smmoth}
\footnotesize
\begin{gathered}
\|\nabla \mathcal{L}_k^{t_1}(W_k^{t_1} ; \boldsymbol{x}, y)-\nabla \mathcal{L}_k^{t_2}(W_k^{t_2} ; \boldsymbol{x}, y)\| \leqslant L_1\|W_k^{t_1}-W_k^{t_2}\|, \\
\forall t_1, t_2>0, k \in\{0,1, \ldots, N-1\},(\boldsymbol{x}, y) \in D_k.
\end{gathered}
\end{equation}
The above formulation can be further derived as:
\begin{equation}
\footnotesize
\mathcal{L}_k^{t_1}-\mathcal{L}_k^{t_2} \leqslant\langle\nabla \mathcal{L}_k^{t_2},(W_k^{t_1}-W_k^{t_2})\rangle+\frac{L_1}{2}\|W_k^{t_1}-W_k^{t_2}\|_2^2 .
\end{equation}
\end{assumption}

\begin{assumption} \label{assump:Unbiased}
\textbf{Unbiased Gradient and Bounded Variance}. Client $k$'s random gradient $g_{W,k}^t=\nabla \mathcal{L}_k^t(W_k^t; \mathcal{B}_k^t)$ ($\mathcal{B}$ is a batch of local data) is unbiased, 
\begin{equation}
\footnotesize
\mathbb{E}_{\mathcal{B}_k^t \subseteq D_k}[g_{W,k}^t]=\nabla \mathcal{L}_k^t(W_k^t),
\end{equation}
and the variance of random gradient $g_{W,k}^t$ is bounded by:
\begin{equation}
\footnotesize
\begin{split}
\mathbb{E}_{\mathcal{B}_k^t \subseteq D_k}[\|\nabla \mathcal{L}_k^t(W_k^t ; \mathcal{B}_k^t)-\nabla \mathcal{L}_k^t(W_k^t)\|_2^2] \leqslant \sigma^2.
\end{split}
\end{equation}    
\end{assumption}

\begin{assumption} \label{assump:BoundedVariation}
\textbf{Bounded Parameter Variation}. The parameter variations of the \homo small feature extractor $\theta_k^t$ and $\theta^t$ before and after aggregation is bounded as
\begin{equation}
\footnotesize
     {\|\theta^t - \theta_k^{t}\|}_2^2 \leq \delta^2.
\end{equation}
\end{assumption} 

Based on the above assumptions, we can derive the following Lemma and Theorem. Detailed proofs are given in Appendix~\ref{app:proof}.

\begin{lemma}\label{lemma:localtraining}
    \textbf{Local Training.} Given Assumptions~\ref{assump:Lipschitz} and \ref{assump:Unbiased}, the loss of an arbitrary client's local model $W$ in the $(t+1)$-th local training round is bounded by 
    \begin{equation}
    \footnotesize
        \mathbb{E}[\mathcal{L}_{(t+1) E}] \leq \mathcal{L}_{t E+0}+(\frac{L_1 \eta^2}{2}-\eta) \sum_{e=0}^E\|\nabla \mathcal{L}_{t E+e}\|_2^2+\frac{L_1 E \eta^2 \sigma^2}{2}. 
    \end{equation}
\end{lemma}

\begin{lemma}\label{lemma:aggregation}
\textbf{Model Aggregation.} Given Assumptions \ref{assump:Unbiased} and \ref{assump:BoundedVariation}, after the $(t+1)$-th local training round, the loss of any client before and after aggregating local \homo small feature extractors at the server is bounded by
\begin{equation}
 \footnotesize   \mathbb{E}\left[\mathcal{L}_{(t+1)E+0}\right]\le\mathbb{E}\left[\mathcal{L}_{tE+1}\right]+{\eta\delta}^2.
\end{equation}
\end{lemma}

\begin{theorem}\label{theorem:one-round}
\textbf{One Complete Round of FL}. Based on Lemma~\ref{lemma:localtraining} and Lemma~\ref{lemma:aggregation}, for any client, after local training, model aggregation, and receiving the new global \homo feature extractor, we get
\begin{equation}
\footnotesize
\mathbb{E}[\mathcal{L}_{(t+1) E+0}] \leq \mathcal{L}_{t E+0}+(\frac{L_1 \eta^2}{2}-\eta) \sum_{e=0}^E\|\nabla \mathcal{L}_{t E+e}\|_2^2+\frac{L_1 E \eta^2 \sigma^2}{2}+\eta \delta^2.
\end{equation}
\end{theorem}

\begin{theorem}\label{theorem:non-convex}
\textbf{Non-convex Convergence Rate of \methodname{}.} Based on Theorem~\ref{theorem:one-round}, for any client and an arbitrary constant $\epsilon>0$, the following holds true:
\begin{equation}
\footnotesize
\begin{aligned}
\frac{1}{T} \sum_{t=0}^{T-1} \sum_{e=0}^{E-1}\|\nabla \mathcal{L}_{t E+e}\|_2^2 &\leq \frac{\frac{1}{T} \sum_{t=0}^{T-1}[\mathcal{L}_{t E+0}-\mathbb{E}[\mathcal{L}_{(t+1) E+0}]]+\frac{L_1 E \eta^2 \sigma^2}{2}+\eta \delta^2}{\eta-\frac{L_1 \eta^2}{2}}<\epsilon, \\
s.t. \   &\eta<\frac{2(\epsilon-\delta^2)}{L_1(\epsilon+E \sigma^2)} .
\end{aligned}
\end{equation}
\end{theorem}
Therefore, we conclude that any client's local model can converge at a non-convex rate $\epsilon \sim \mathcal{O}(\frac{1}{T})$ under \methodname{}.

\section{Experimental Evaluation}

To evaluate the effectiveness of \methodname{}, we implement it and $7$ state-of-the-art baselines by Pytorch and compare them over 2 benchmark datasets on $4$ NVIDIA GeForce RTX 3090 GPUs.

\subsection{Experiment Setup}

\textbf{Datasets.} We evaluate \methodname{} and baselines on CIFAR-10 and CIFAR-100 \footnote{\url{https://www.cs.toronto.edu/\%7Ekriz/cifar.html}} \cite{cifar} image classification benchmark datasets. CIFAR-10 comprises $6,000$ $32\times32$ color images across $10$ classes, with $5,000$ images in the training set and $1,000$ images in the testing set. CIFAR-100 contains $100$ classes of color images, each with $500$ training images and $100$ testing images. 
To construct non-IID datasets, we adopt two data partitioning strategies: (1) \textbf{Pathological}: Following \cite{pFedHN}, we allocate $2$ classes to each client on CIFAR-10 and use Dirichlet distribution to generate varying counts of the same class for different clients, denoted as (non-IID: 2/10). We assign $10$ classes to each client on CIFAR-100, marked as (non-IID: 10/100). (2) \textbf{Practical}: Following \citet{FedAPEN}, we allocate all classes to each client and utilize Dirichlet distribution($\gamma$) to control the proportions of each class across clients.
After non-IID division, each client's local dataset is divided into training and testing sets in an $8:2$ ratio, ensuring both sets follow the same distribution.

\textbf{Base Models.} We assess \methodname{} and baselines in both model-\homo and model-\hetero FL scenarios. For model-\homo settings, all clients hold the same CNN-1 shown in Table~\ref{tab:model-structures} (Appendix~\ref{app:experiment}). In model-\hetero settings, $5$ \hetero CNN models are evenly allocated to different clients, with assignment IDs determined by client ID modulo $5$.

\textbf{Comparison Baselines.} We compare \methodname{} against \sota MHPFL algorithms from the three most relevant categories of public data-independent MHPFL algorithms outlined in Section~\ref{sec:related}.
\begin{itemize}
    \item {\tt{Standalone.}} Clients only utilize their local data to train models without FL process.
    \item \textbf{MHPFL by Knowledge Distillation:} {\tt{FD}}~\citep{FD}, {\tt{FedProto}}~\citep{FedProto}.
    \item \textbf{MHPFL by Model Mixture:} {\tt{LG-FedAvg}}~\citep{LG-FedAvg}.
    \item \textbf{MHPFL by Mutual Learning:} {\tt{FML}}~\citep{FML}, {\tt{FedKD}}~\citep{FedKD}, and the latest {\tt{FedAPEN}}~\citep{FedAPEN}. \methodname{} belongs to this category.
\end{itemize}

\textbf{Evaluation Metrics.} We measure the model performance, communication cost, and computational overhead of all algorithms.
\begin{itemize}
    \item \textbf{Model Performance.} We evaluate each client's local model's \textbf{individual test accuracy} (\%) on the local testing set and calculate their \textbf{mean test accuracy}.
    \item \textbf{Communication Cost.} We monitor the communication rounds required to reach \textbf{target mean accuracy} and quantify communication costs by multiplying rounds with the mean parameter capacity transmitted in one round.
    \item \textbf{Computation Overhead.} We calculate computation overhead by multiplying the communication rounds required to achieve \textbf{target mean accuracy} with the local mean computational FLOPs in one round.
\end{itemize}

\textbf{Training Strategy.}
We conduct a grid search to identify the optimal FL settings and specific hyperparameters for all algorithms. In FL settings, we evaluate all algorithms over $T=\{100, 500\}$ communication rounds, $\{1, 10\}$ local epochs, $\{64, 128, 256, 512\}$ batch sizes, and the SGD optimizer with learning rate $\{0.001, 0.01, 0.1, 1\}$. For \methodname{}, the \homo small feature extractor and the \hetero large model have the same learning rate (\emph{i.e.}, $\eta_\theta=\eta_\omega$). We report the highest accuracy achieved by all algorithms.

\subsection{Results and Discussion}
To test algorithms in different FL scenarios with diverse number of clients $N$ and client participation rates $C$, we design three settings: $\{(N=10, C=100\%), (N=50, C=20\%), (N=100, C=10\%)\}$.

\subsubsection{Model Homogeneity}
Table~\ref{tab:compare-homo} shows that \textit{\methodname{} consistently achieves the highest accuracy}, surpassing each setting's \sota baseline ({{\tt{LG-FedAvg}}, {\tt{Standalone}}, {\tt{Standalone}},  {\tt{FedProto}}, {\tt{Standalone}}, {\tt{Standalone}}}) by up to $1.74\%$, and improving the accuracy by up to $5.47\%$ compared with the same-category best baseline ({{\tt{FedAPEN}}, {\tt{FedAPEN}}, {\tt{FedAPEN}}, {\tt{FedAPEN}}, {\tt{FedKD}}, {\tt{FedKD}}}). The results indicate that \methodname{} efficiently boosts model accuracy through adaptive data-level personalization.

\subsubsection{Model Heterogeneity}
In this scenario, \methodname{} and other mutual learning-based MHPFL baselines utilize the smallest CNN-5 (Table~\ref{tab:model-structures}, Appendix~\ref{app:experiment}) as \homo feature extractors or models.

\textbf{Mean Accuracy.}
Table~\ref{tab:compare-hetero} shows that \textit{\methodname{} consistently outperforms other baselines}, improving test accuracy by up to $2.80\%$ compared to the \sota baseline under each setting ({{\tt{Standalone}}, {\tt{FedProto}}, {\tt{FedProto}}, {\tt{Standalone}}, {\tt{FedProto}}, {\tt{FedProto}}}). It improves test accuracy by up to $22.16\%$ compared to the same-category best baseline ({\tt{FedKD}}). Figure~\ref{fig:compare-hetero-converge} (Appendix~\ref{app:experiment}) shows 
that \textit{\methodname{} achieves faster convergence and higher model accuracy} across most FL settings, particularly noticeable on CIFAR-100.

\begin{figure}[t]
\centering
\begin{minipage}[t]{0.5\linewidth}
\centering
\includegraphics[width=1.7in]{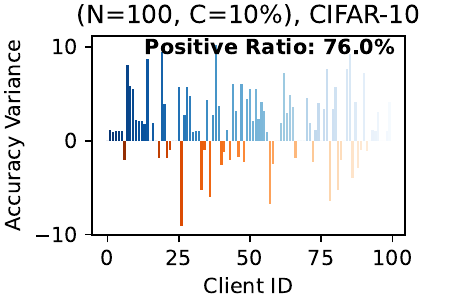}
\end{minipage}%
\begin{minipage}[t]{0.5\linewidth}
\centering
\includegraphics[width=1.7in]{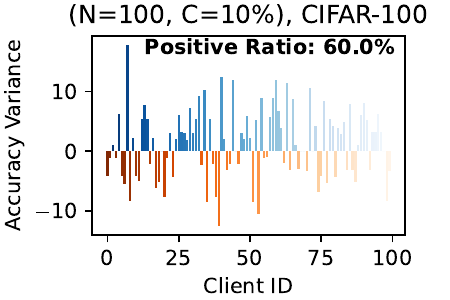}
\end{minipage}%
\vspace{-1em}
\caption{Accuracy distribution for individual clients.}
\label{fig:compare-individual}
\vspace{-1em}
\end{figure}

\textbf{Individual Accuracy.} Figure~\ref{fig:compare-individual} shows the performance variance of \methodname{} and the \sota baseline - {\tt{FedProto}} in terms of the individual accuracy of each client under ($N=100, C=10\%$). Most clients (CIFAR-10: $76\%$, CIFAR-100: $60\%$) with \methodname{} achieve higher accuracy than {\tt{FedProto}}. This demonstrates that \methodname{} with data-level \persN dynamically adapts to local data distribution and learns more \gen and \pers knowledge from local data.

\begin{table}[t]
\centering
\caption{Mean accuracy in \textit{model-\homo} FL scenarios.}
\vspace{-1em}
\resizebox{\linewidth}{!}{%
\begin{tabular}{|l|cc|cc|cc|}
\hline
FL Setting                    & \multicolumn{2}{c|}{N=10, C=100\%}                                                                                                                   & \multicolumn{2}{c|}{N=50, C=20\%}                                                                                                                    & \multicolumn{2}{c|}{N=100, C=10\%}                                                                                                                   \\ \hline
Method                        & \multicolumn{1}{c|}{CIFAR-10}                                                      & CIFAR-100                                                     & \multicolumn{1}{c|}{CIFAR-10}                                                      & CIFAR-100                                                     & \multicolumn{1}{c|}{CIFAR-10}                                                      & CIFAR-100                                                     \\ \hline
Standalone                    & \multicolumn{1}{c|}{96.35}                                                         & \cellcolor[HTML]{C0C0C0}\textbf{74.32}                        & \multicolumn{1}{c|}{\cellcolor[HTML]{C0C0C0}\textbf{95.25}}                        & 62.38                                                         & \multicolumn{1}{c|}{\cellcolor[HTML]{C0C0C0}\textbf{92.58}}                        & \cellcolor[HTML]{C0C0C0}\textbf{54.93}                        \\
LG-FedAvg~\citep{LG-FedAvg}                     & \multicolumn{1}{c|}{\cellcolor[HTML]{C0C0C0}\textbf{96.47}}                        & 73.43                                                         & \multicolumn{1}{c|}{94.20}                                                         & 61.77                                                         & \multicolumn{1}{c|}{90.25}                                                         & 46.64                                                         \\
FD~\citep{FD}                            & \multicolumn{1}{c|}{96.30}                                                         & -                                                             & \multicolumn{1}{c|}{-}                                                             & -                                                             & \multicolumn{1}{c|}{-}                                                             & -                                                             \\
FedProto~\citep{FedProto}                      & \multicolumn{1}{c|}{95.83}                                                         & 72.79                                                         & \multicolumn{1}{c|}{95.10}                                                         & \cellcolor[HTML]{C0C0C0}\textbf{62.55}                        & \multicolumn{1}{c|}{91.19}                                                         & 54.01                                                         \\ \hline
FML~\citep{FML}                           & \multicolumn{1}{c|}{94.83}                                                         & 70.02                                                         & \multicolumn{1}{c|}{93.18}                                                         & 57.56                                                         & \multicolumn{1}{c|}{87.93}                                                         & 46.20                                                         \\
FedKD~\citep{FedKD}                         & \multicolumn{1}{c|}{94.77}                                                         & 70.04                                                         & \multicolumn{1}{c|}{92.93}                                                         & 57.56                                                         & \multicolumn{1}{c|}{\cellcolor[HTML]{EFEFEF}\textbf{90.23}}                        & \cellcolor[HTML]{EFEFEF}\textbf{50.99}                        \\
FedAPEN~\citep{FedAPEN}                       & \multicolumn{1}{c|}{\cellcolor[HTML]{EFEFEF}\textbf{95.38}}                        & \cellcolor[HTML]{EFEFEF}\textbf{71.48}                        & \multicolumn{1}{c|}{\cellcolor[HTML]{EFEFEF}\textbf{93.31}}                        & \cellcolor[HTML]{EFEFEF}\textbf{57.62}                        & \multicolumn{1}{c|}{87.97}                                                         & 46.85                                                         \\ \hline
\textbf{\methodname{}}               & \multicolumn{1}{c|}{\cellcolor[HTML]{9B9B9B}{\color[HTML]{000000} \textbf{96.80}}} & \cellcolor[HTML]{9B9B9B}{\color[HTML]{000000} \textbf{76.06}} & \multicolumn{1}{c|}{\cellcolor[HTML]{9B9B9B}{\color[HTML]{000000} \textbf{95.80}}} & \cellcolor[HTML]{9B9B9B}{\color[HTML]{000000} \textbf{63.06}} & \multicolumn{1}{c|}{\cellcolor[HTML]{9B9B9B}{\color[HTML]{000000} \textbf{93.55}}} & \cellcolor[HTML]{9B9B9B}{\color[HTML]{000000} \textbf{56.46}} \\ \hline
\rowcolor[HTML]{EFEFEF} 
\textit{\methodname{}-Best B.}       & \multicolumn{1}{c|}{\cellcolor[HTML]{EFEFEF}{\color[HTML]{000000} \textit{0.33}}}  & {\color[HTML]{000000} \ul \textit{1.74}}                          & \multicolumn{1}{c|}{\cellcolor[HTML]{EFEFEF}{\color[HTML]{000000} \textit{0.55}}}  & {\color[HTML]{000000} \textit{0.51}}                          & \multicolumn{1}{c|}{\cellcolor[HTML]{EFEFEF}{\color[HTML]{000000} \textit{0.97}}}  & {\color[HTML]{000000} \textit{1.53}}                          \\
\rowcolor[HTML]{EFEFEF} 
\textit{\methodname{}-Best S.C.B.} & \multicolumn{1}{c|}{\cellcolor[HTML]{EFEFEF}{\color[HTML]{000000} \textit{1.42}}}  & {\color[HTML]{000000} \textit{4.58}}                          & \multicolumn{1}{c|}{\cellcolor[HTML]{EFEFEF}{\color[HTML]{000000} \textit{2.49}}}  & {\color[HTML]{000000} \textit{5.44}}                          & \multicolumn{1}{c|}{\cellcolor[HTML]{EFEFEF}{\color[HTML]{000000} \textit{3.32}}}  & {\color[HTML]{000000} \ul \textit{5.47}}                          \\ \hline
\end{tabular}
}
\raggedright
\footnotesize Note: ``-'' denotes failure to converge. ``Best B.'' indicates the best baseline. ``Best S.C.B.'' means the best same-category baseline.
\label{tab:compare-homo}
\vspace{-1em}
\end{table}

\begin{table}[t]
\centering
\caption{Mean accuracy in \textit{model-\hetero} FL scenarios.}
\vspace{-1em}
\resizebox{\linewidth}{!}{%
\begin{tabular}{|l|cc|cc|cc|}
\hline
FL Setting                    & \multicolumn{2}{c|}{N=10, C=100\%}                                                                                                                   & \multicolumn{2}{c|}{N=50, C=20\%}                                                                                                                    & \multicolumn{2}{c|}{N=100, C=10\%}                                                                                                                   \\ \hline
Method                        & \multicolumn{1}{c|}{CIFAR-10}                                                      & CIFAR-100                                                     & \multicolumn{1}{c|}{CIFAR-10}                                                      & CIFAR-100                                                     & \multicolumn{1}{c|}{CIFAR-10}                                                      & CIFAR-100                                                     \\ \hline
Standalone                    & \multicolumn{1}{c|}{\cellcolor[HTML]{C0C0C0}{\color[HTML]{000000} \textbf{96.53}}} & {\color[HTML]{000000} 72.53}                                  & \multicolumn{1}{c|}{{\color[HTML]{000000} 95.14}}                                  & \cellcolor[HTML]{C0C0C0}{\color[HTML]{000000} \textbf{62.71}} & \multicolumn{1}{c|}{{\color[HTML]{000000} 91.97}}                                  & {\color[HTML]{000000} 53.04}                                  \\
LG-FedAvg~\citep{LG-FedAvg}                     & \multicolumn{1}{c|}{{\color[HTML]{000000} 96.30}}                                  & {\color[HTML]{000000} 72.20}                                  & \multicolumn{1}{c|}{{\color[HTML]{000000} 94.83}}                                  & {\color[HTML]{000000} 60.95}                                  & \multicolumn{1}{c|}{{\color[HTML]{000000} 91.27}}                                  & {\color[HTML]{000000} 45.83}                                  \\
FD~\citep{FD}                            & \multicolumn{1}{c|}{{\color[HTML]{000000} 96.21}}                                  & {\color[HTML]{000000} -}                                      & \multicolumn{1}{c|}{{\color[HTML]{000000} -}}                                      & {\color[HTML]{000000} -}                                      & \multicolumn{1}{c|}{{\color[HTML]{000000} -}}                                      & {\color[HTML]{000000} -}                                      \\
FedProto~\citep{FedProto}                      & \multicolumn{1}{c|}{{\color[HTML]{000000} 96.51}}                                  & \cellcolor[HTML]{C0C0C0}{\color[HTML]{000000} \textbf{72.59}} & \multicolumn{1}{c|}{\cellcolor[HTML]{C0C0C0}{\color[HTML]{000000} \textbf{95.48}}} & {\color[HTML]{000000} 62.69}                                  & \multicolumn{1}{c|}{\cellcolor[HTML]{C0C0C0}{\color[HTML]{000000} \textbf{92.49}}} & \cellcolor[HTML]{C0C0C0}{\color[HTML]{000000} \textbf{53.67}} \\ \hline
FML~\citep{FML}                           & \multicolumn{1}{c|}{{\color[HTML]{000000} 30.48}}                                  & {\color[HTML]{000000} 16.84}                                  & \multicolumn{1}{c|}{{\color[HTML]{000000} -}}                                      & {\color[HTML]{000000} 21.96}                                  & \multicolumn{1}{c|}{{\color[HTML]{000000} -}}                                      & {\color[HTML]{000000} 15.21}                                  \\
FedKD~\citep{FedKD}                         & \multicolumn{1}{c|}{\cellcolor[HTML]{EFEFEF}{\color[HTML]{000000} \textbf{80.20}}} & \cellcolor[HTML]{EFEFEF}{\color[HTML]{000000} \textbf{53.23}} & \multicolumn{1}{c|}{\cellcolor[HTML]{EFEFEF}{\color[HTML]{000000} \textbf{77.37}}} & \cellcolor[HTML]{EFEFEF}{\color[HTML]{000000} \textbf{44.27}} & \multicolumn{1}{c|}{\cellcolor[HTML]{EFEFEF}{\color[HTML]{000000} \textbf{73.21}}} & \cellcolor[HTML]{EFEFEF}{\color[HTML]{000000} \textbf{37.21}} \\
FedAPEN~\citep{FedAPEN}                       & \multicolumn{1}{c|}{{\color[HTML]{000000} \textbf{-}}}                             & {\color[HTML]{000000} \textbf{-}}                             & \multicolumn{1}{c|}{{\color[HTML]{000000} \textbf{-}}}                             & {\color[HTML]{000000} \textbf{-}}                             & \multicolumn{1}{c|}{{\color[HTML]{000000} -}}                                      & {\color[HTML]{000000} -}                                      \\ \hline
\textbf{\methodname{}}               & \multicolumn{1}{c|}{\cellcolor[HTML]{9B9B9B}{\color[HTML]{000000} \textbf{96.58}}} & \cellcolor[HTML]{9B9B9B}{\color[HTML]{000000} \textbf{75.39}} & \multicolumn{1}{c|}{\cellcolor[HTML]{9B9B9B}{\color[HTML]{000000} \textbf{95.84}}} & \cellcolor[HTML]{9B9B9B}{\color[HTML]{000000} \textbf{63.30}} & \multicolumn{1}{c|}{\cellcolor[HTML]{9B9B9B}{\color[HTML]{000000} \textbf{93.07}}} & \cellcolor[HTML]{9B9B9B}{\color[HTML]{000000} \textbf{54.78}} \\ \hline
\rowcolor[HTML]{EFEFEF} 
\textit{\methodname{}-Best B.}       & \multicolumn{1}{c|}{\cellcolor[HTML]{EFEFEF}{\color[HTML]{000000} \textit{0.05}}}  & {\color[HTML]{000000} {\ul \textit{2.80}}}                    & \multicolumn{1}{c|}{\cellcolor[HTML]{EFEFEF}{\color[HTML]{000000} \textit{0.36}}}  & {\color[HTML]{000000} \textit{0.59}}                          & \multicolumn{1}{c|}{\cellcolor[HTML]{EFEFEF}{\color[HTML]{000000} \textit{0.58}}}  & {\color[HTML]{000000} \textit{1.11}}                          \\
\rowcolor[HTML]{EFEFEF} 
\textit{\methodname{}-Best S.C.B.} & \multicolumn{1}{c|}{\cellcolor[HTML]{EFEFEF}{\color[HTML]{000000} \textit{16.38}}} & {\color[HTML]{000000} {\ul \textit{22.16}}}                   & \multicolumn{1}{c|}{\cellcolor[HTML]{EFEFEF}{\color[HTML]{000000} \textit{18.47}}} & {\color[HTML]{000000} \textit{19.03}}                         & \multicolumn{1}{c|}{\cellcolor[HTML]{EFEFEF}{\color[HTML]{000000} \textit{19.86}}} & {\color[HTML]{000000} \textit{17.57}}                         \\ \hline
\end{tabular}
}
\raggedright
\footnotesize Note: ``-'' denotes failure to converge. ``Best B.'' indicates the best baseline. ``Best S.C.B.'' means the best same-category baseline.
\label{tab:compare-hetero}
\vspace{-1em}
\end{table}

\begin{figure}[t]
\centering
\begin{minipage}[t]{0.5\linewidth}
\centering
\includegraphics[width=1.6in]{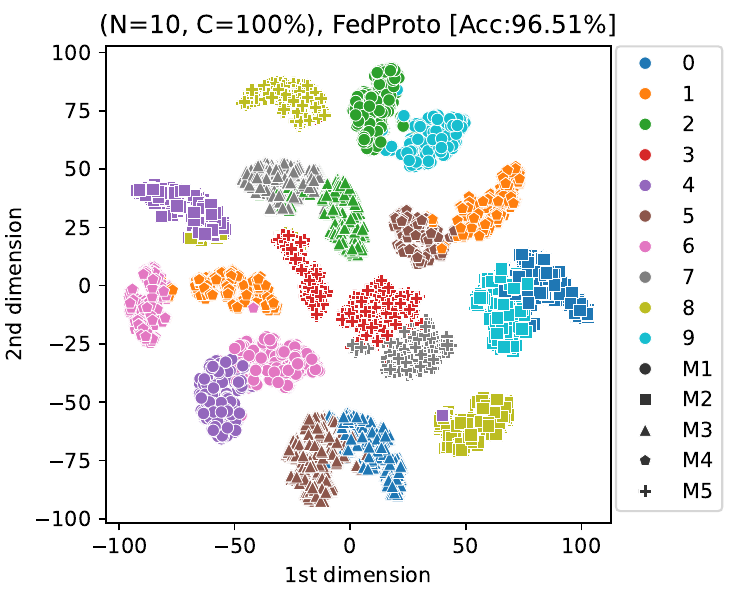}
\end{minipage}%
\begin{minipage}[t]{0.5\linewidth}
\centering
\includegraphics[width=1.6in]{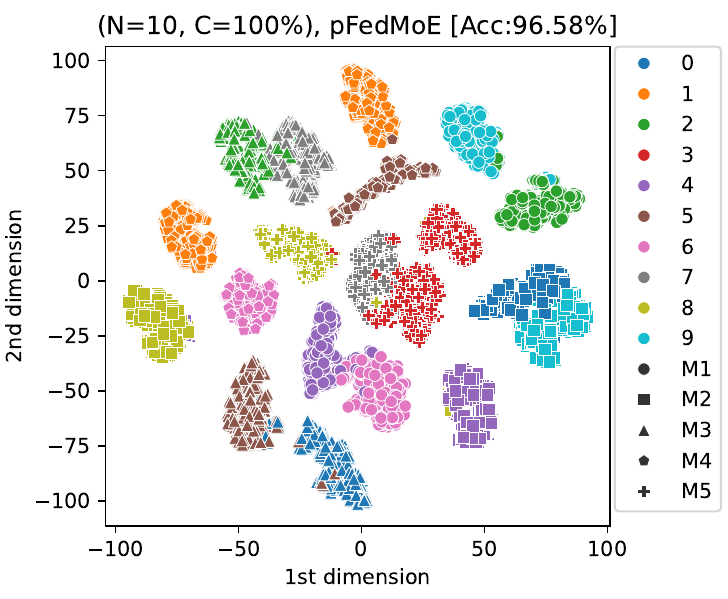}
\end{minipage}%
\vspace{-1em}
\caption{T-SNE representation visualization results for {\tt{FedProto}} and \methodname{} on CIFAR-10 (non-IID: 2/10).}
\label{fig:compare-TSNE}
\vspace{-1em}
\end{figure}

\textbf{Personalization Analysis.} To explore the \persN level of \methodname{} and {\tt{FedProto}}, we extract the \rep of each local data sample from each client produced by \methodname{}'s local \hetero MoE and {\tt{FedProto}}'s local \hetero feature extractor on CIFAR-10 (non-IID: 2/10) under ($N=100, C=10\%$). We adopt \textit{T-SNE}~\citep{TSNE-JMLR} to compress extracted \reps to $2$-dimension vectors and visualize them in Figure~\ref{fig:compare-TSNE}. Limited by plotting spaces, more clients $N=\{50, 100\}$ and CIFAR-100 with $100$ classes cannot be clearly depicted. Figure~\ref{fig:compare-TSNE} shows that \reps from one client's $2$ seen classes are close together, while \reps from different clients are further apart, indicating that two algorithms indeed yield \pers local \hetero models. Regarding \reps of one client's $2$ classes, they present ``intra-class compactness, inter-class separation'', signifying a strong classification capability. Notably, representations of one client's $2$ classes under \methodname{} exhibit clearer decision boundaries (\emph{i.e.}, \reps within the same cluster are more compact), suggesting high classification performance and \persN.

\begin{figure}[!t]
\centering
\begin{minipage}[t]{\linewidth}
\centering
\includegraphics[width=3in]{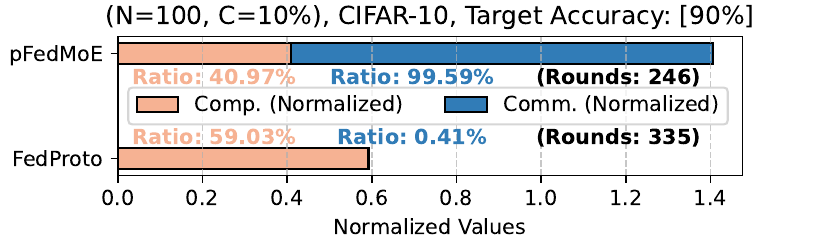}
\end{minipage}%

\begin{minipage}[t]{\linewidth}
\centering
\includegraphics[width=3in]{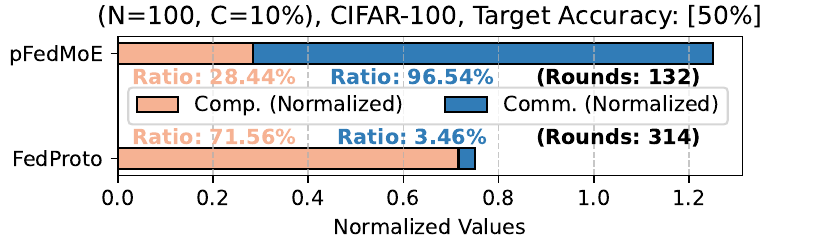}
\end{minipage}%
\vspace{-12pt}
\caption{Computation (Comp.), communication (Comm.), and rounds required for reaching target mean accuracy.}
\label{fig:compare-comm-comp}
\vspace{-12pt}
\end{figure}

\textbf{Overhead Analysis.} Figure~\ref{fig:compare-comm-comp} shows the required number of rounds, communication, and computation costs for \methodname{} and {\tt{FedProto}} to reach $90\%$ and $50\%$ test accuracy under the most complex ($N=100, C=10\%$) setting.
For fair comparisons, we normalize communication and computation costs, considering only their different dimensions (i.e., number of parameters, FLOPs).

\textbf{Computation}. \textit{\methodname{} incurs lower computational costs than {\tt{FedProto}}}. This is because {\tt{FedProto}} requires extracting \reps for all local data samples after training local \hetero models, while \methodname{} trains an MoE and the prediction header simultaneously. One round of computation in \methodname{} is less than {\tt{FedProto}}. Since \methodname{} also requires fewer rounds to reach the target accuracy, it incurs lower total computation costs. 

\textbf{Communication}. \methodname{} incurs higher communication costs than {\tt{FedProto}}. This is because in one round, clients with {\tt{FedProto}} transmit seen-class \reps to the server, while clients with \methodname{} transmit \homo small feature extractors. This, the former consumes lower communication costs per round. Despite requiring fewer rounds to reach the target accuracy, \methodname{} still consumes higher total communication costs. However, compared with transmitting complete local models in {\tt{FedAvg}}, \methodname{} still incurs lower communication overheads. Therefore, \textit{\methodname{} achieves highly efficient computation with acceptable communication costs, while delivering superior model accuracy.}

\subsection{Case Studies}

\subsubsection{Robustness to Pathological Non-IIDness}
\begin{figure}[t]
\centering
\begin{minipage}[t]{0.5\linewidth}
\centering
\includegraphics[width=1.75in]{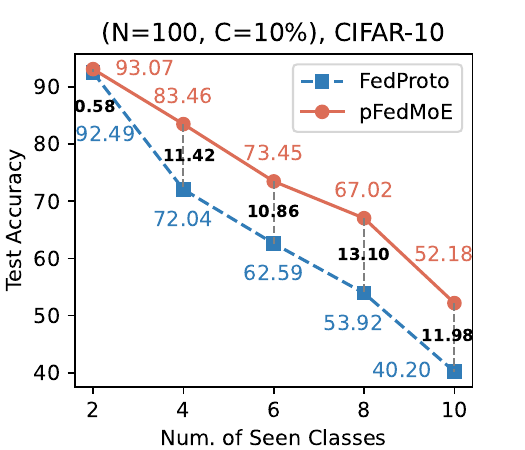}
\end{minipage}%
\begin{minipage}[t]{0.5\linewidth}
\centering
\includegraphics[width=1.75in]{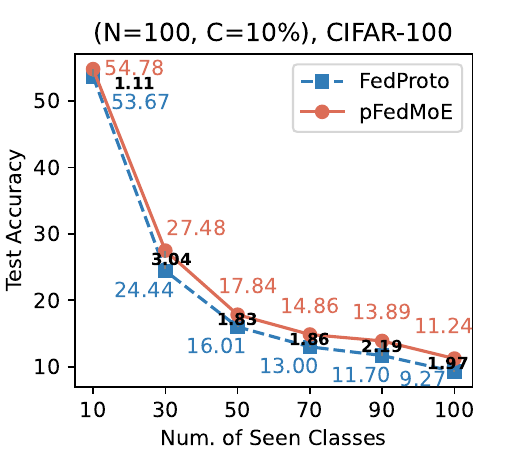}
\end{minipage}%
\vspace{-1em}
\caption{Robustness to \textit{pathological} non-IIDness.}
\label{fig:case-noniid-class}
\vspace{-1em}
\end{figure}

In model \hetero FL with ($N=100, C=10\%$), the number of seen classes assigned to one client varies as $\{2, 4, 6, 8, 10\}$ on CIFAR-10 and $\{10, 30, 50, 70, 90, 100\}$ on CIFAR-100. 
Figure~\ref{fig:case-noniid-class} shows that model accuracy drops as non-IIDness reduces (number of seen classes rises), as clients with more classes exhibit degraded classification ability for each class (\emph{i.e.}, model \genN improves, but \persN drops). 
Besides, \methodname{} consistently achieves higher test accuracy than {\tt{FedProto}} across various non-IIDnesses settings, indicating its robustness to pathological non-IIDness. 
Moreover, \textit{\methodname{} achieves higher accuracy improvements compared to {\tt{FedProto}} under IID settings than under non-IID settings}, \emph{e.g.}, $+13.10\%$ on CIFAR-10 (non-IID: 8/10) and $+3.04\%$ on CIFAR-100 (non-IID: 30/100). This suggests that {\tt{FedProto}} is more effective under non-IID settings than under IID settings, consistent with the behavior of most \pers FL algorithms \citep{FedAPEN}. In contrast, \methodname{} adapts to both IID and non-IID settings by the \pers gating network to dynamically balance global \genN and local personalization.

\begin{figure}[t]
\centering
\begin{minipage}[t]{0.5\linewidth}
\centering
\includegraphics[width=1.75in]{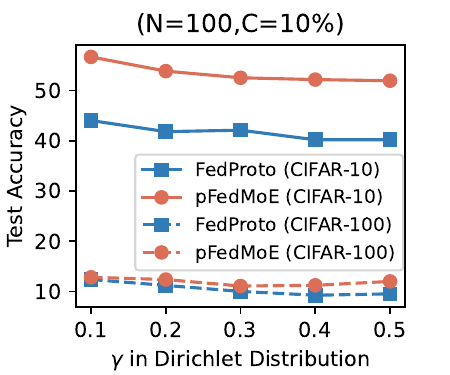}
\end{minipage}%
\begin{minipage}[t]{0.5\linewidth}
\centering
\includegraphics[width=1.75in]{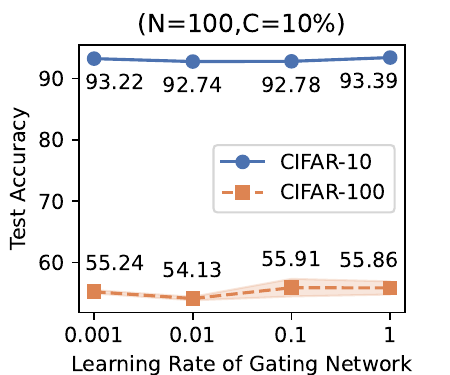}
\end{minipage}%
\vspace{-1em}
\caption{Left: robustness to \textit{practical} non-IIDness. Right: sensitivity to the gating network learning rate.}
\label{fig:case-noniid-distribution-sensitivity}
\vspace{-1em}
\end{figure}

\subsubsection{Robustness to Practical Non-IIDness}
In model \hetero FL with ($N=100, C=10\%$), $\gamma$ of the Dirichlet distribution varies as $\{0.1, 0.2, 0.3, 0.4, 0.5\}$. Figure~\ref{fig:case-noniid-distribution-sensitivity}(left) shows that \methodname{} consistently achieves higher accuracy than {\tt{FedProto}}, indicating its robustness to practical non-IIDness. Similar to Figure~\ref{fig:case-noniid-class}, model accuracy drops as non-IIDness reduces ($\gamma$ rises). \methodname{} improves test accuracy more under IID settings than under non-IID settings.

\subsubsection{Sensitivity Analysis}
Only one extra hyperparameter, the learning rate $\eta_\varphi$ of the gating network, is introduced by \methodname{}. In model \hetero FL with ($N=100, C=10\%$), we evaluate \methodname{} with $\eta_\varphi=\{0.001,0.01,0.1,1\}$ on CIFAR-10 (non-IID: 2/10) and CIFAR-100 (non-IID: 10/100). We select three random seeds to execute $3$ trails for each test. Figure~\ref{fig:case-noniid-distribution-sensitivity} (right) shows the accuracy mean (dots) and variation (shadow). \methodname{} achieves stable accuracy across various gating network learning rates, indicating that it is not sensitive to $\eta_\varphi$.

\subsubsection{Weight Analysis} 
We analyze \methodname{}'s gating network output weights in a model \hetero FL with ($N=100, C=10\%$) on CIFAR-10 (non-IID: 2/10) and CIFAR-100 (non-IID: 10/100).

\begin{figure}[t]
\centering
\begin{minipage}[t]{0.5\linewidth}
\centering
\includegraphics[width=1.75in]{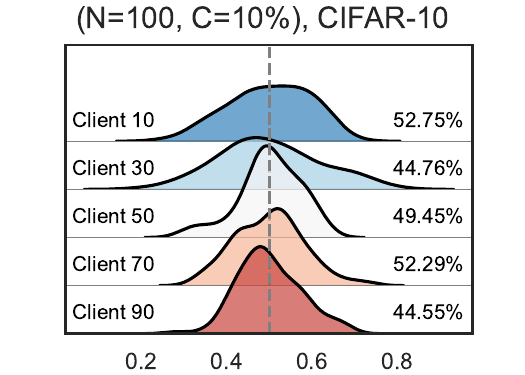}
\end{minipage}%
\begin{minipage}[t]{0.5\linewidth}
\centering
\includegraphics[width=1.75in]{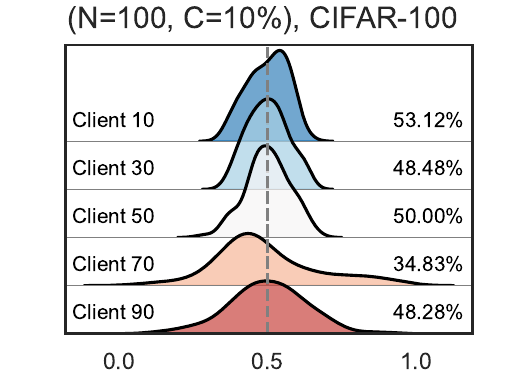}
\end{minipage}%
\caption{Produced weight distributions vary as \textit{clients}.}
\label{fig:compare-weights_clients}
\end{figure}

\textbf{Client Perspective.} We randomly select $5$ clients and visualize the probability distribution of the weights produced by the final local gating network for the local \pers \hetero large feature extractor on all local testing data. 
Figure~\ref{fig:compare-weights_clients} shows that different clients with non-IID data exhibit diverse weight distributions, indicating that the weights produced by the local \pers gating network for different clients are indeed \pers to local data distributions. Besides, most weights are concentrated around $50\%$, with some exceeding $50\%$, suggesting that the \gen features extracted by the small \homo feature extractor and the \pers features extracted by the large \hetero feature extractor contribute comparably to model performance. The \pers output weights dynamically balance \genN and personalization.

\begin{figure}[t]
\centering
\begin{minipage}[t]{0.5\linewidth}
\centering
\includegraphics[width=1.75in]{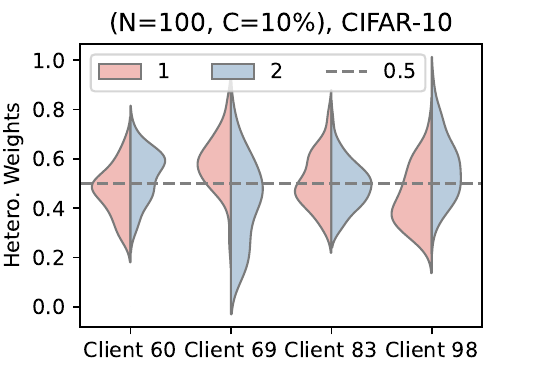}
\end{minipage}%
\begin{minipage}[t]{0.5\linewidth}
\centering
\includegraphics[width=1.75in]{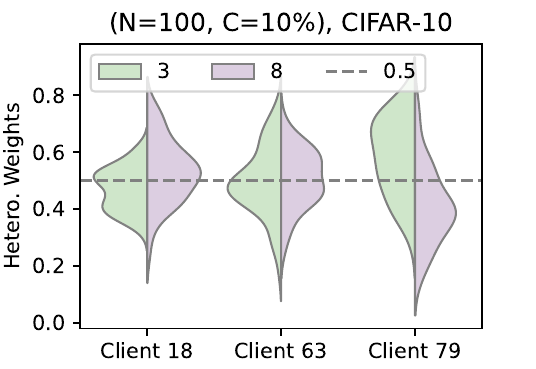}
\end{minipage}%
\caption{Produced weight distributions vary as \textit{classes}.}
\label{fig:compare-weights_2class}
\end{figure}

\textbf{Class Perspective.} Upon identifying client sets with the same seen classes,
we find $4$ clients sharing classes $[1, 2]$, and $3$ clients sharing classes $[3, 8]$ on CIFAR-10. However, no pair of clients have overlapping seen classes on CIFAR-100 as $100$ classes are assigned to $100$ clients. Figure~\ref{fig:compare-weights_2class} shows that one class across different clients performs various weight distributions and different classes within one client also exhibit diverse weight distributions, highlighting that \methodname{} indeed achieves data-level personalization.

\section{Conclusions and Future Work}
In this paper, we proposed a novel model-\hetero \pers federated learning algorithm, \methodname{}, to achieve data-level \persN by leveraging a mixture of experts (MoE). Each client's local complete model consists of a \hetero MoE (a share \homo small feature extractor (global expert), a local \hetero large model's feature extractor (local expert), a local \pers gating network) and a local \hetero large model's \pers prediction header.
\methodname{} exchanges knowledge from local \hetero models across different clients by sharing \homo small feature extractors, and it achieves data-level \persN adaptive to local non-IID data distribution by MoE to balance \genN and personalization. 
Theoretical analysis proves its $\mathcal{O}(1/T)$ non-convex convergence rate.
Extensive experiments demonstrate that \methodname{} obtains the \sota model accuracy with lower computation overheads and acceptable communication costs. 

In subsequent research, we will explore how \methodname{} performs in federated continuous learning (FCL) scenarios involving steaming data with distribution drift.


\bibliographystyle{ACM-Reference-Format}
\bibliography{sample-base}


\begin{thebibliography}{58}


\ifx \showCODEN    \undefined \def \showCODEN     #1{\unskip}     \fi
\ifx \showDOI      \undefined \def \showDOI       #1{#1}\fi
\ifx \showISBNx    \undefined \def \showISBNx     #1{\unskip}     \fi
\ifx \showISBNxiii \undefined \def \showISBNxiii  #1{\unskip}     \fi
\ifx \showISSN     \undefined \def \showISSN      #1{\unskip}     \fi
\ifx \showLCCN     \undefined \def \showLCCN      #1{\unskip}     \fi
\ifx \shownote     \undefined \def \shownote      #1{#1}          \fi
\ifx \showarticletitle \undefined \def \showarticletitle #1{#1}   \fi
\ifx \showURL      \undefined \def \showURL       {\relax}        \fi
\providecommand\bibfield[2]{#2}
\providecommand\bibinfo[2]{#2}
\providecommand\natexlab[1]{#1}
\providecommand\showeprint[2][]{arXiv:#2}

\bibitem[Ahn et~al\mbox{.}(2019)]%
        {HFD1}
\bibfield{author}{\bibinfo{person}{Jin{-}Hyun Ahn} {et~al\mbox{.}}} \bibinfo{year}{2019}\natexlab{}.
\newblock \showarticletitle{Wireless Federated Distillation for Distributed Edge Learning with Heterogeneous Data}. In \bibinfo{booktitle}{\emph{Proc. {PIMRC}}}. \bibinfo{publisher}{{IEEE}}, \bibinfo{address}{Istanbul, Turkey}, \bibinfo{pages}{1--6}.
\newblock


\bibitem[Ahn et~al\mbox{.}(2020)]%
        {HFD2}
\bibfield{author}{\bibinfo{person}{Jin{-}Hyun Ahn} {et~al\mbox{.}}} \bibinfo{year}{2020}\natexlab{}.
\newblock \showarticletitle{Cooperative Learning {VIA} Federated Distillation {OVER} Fading Channels}. In \bibinfo{booktitle}{\emph{Proc. {ICASSP}}}. \bibinfo{publisher}{{IEEE}}, \bibinfo{address}{Barcelona, Spain}, \bibinfo{pages}{8856--8860}.
\newblock


\bibitem[Alam et~al\mbox{.}(2022)]%
        {FedRolex}
\bibfield{author}{\bibinfo{person}{Samiul Alam} {et~al\mbox{.}}} \bibinfo{year}{2022}\natexlab{}.
\newblock \showarticletitle{FedRolex: Model-Heterogeneous Federated Learning with Rolling Sub-Model Extraction}. In \bibinfo{booktitle}{\emph{Proc. NeurIPS}}. \bibinfo{publisher}{{}}, \bibinfo{address}{virtual}.
\newblock


\bibitem[Babakniya et~al\mbox{.}(2023)]%
        {FLASH}
\bibfield{author}{\bibinfo{person}{Sara Babakniya} {et~al\mbox{.}}} \bibinfo{year}{2023}\natexlab{}.
\newblock \showarticletitle{Revisiting Sparsity Hunting in Federated Learning: Why does Sparsity Consensus Matter?}
\newblock \bibinfo{journal}{\emph{Transactions on Machine Learning Research}} (\bibinfo{year}{2023}).
\newblock
\showISSN{2835-8856}
\urldef\tempurl%
\url{https://openreview.net/forum?id=iHyhdpsnyi}
\showURL{%
\tempurl}


\bibitem[Chang et~al\mbox{.}(2021)]%
        {Cronus}
\bibfield{author}{\bibinfo{person}{Hongyan Chang} {et~al\mbox{.}}} \bibinfo{year}{2021}\natexlab{}.
\newblock \showarticletitle{Cronus: Robust and Heterogeneous Collaborative Learning with Black-Box Knowledge Transfer}. In \bibinfo{booktitle}{\emph{Proc. NeurIPS Workshop}}. \bibinfo{publisher}{{}}, \bibinfo{address}{virtual}.
\newblock


\bibitem[Chen et~al\mbox{.}(2023)]%
        {pFedGate}
\bibfield{author}{\bibinfo{person}{Daoyuan Chen} {et~al\mbox{.}}} \bibinfo{year}{2023}\natexlab{}.
\newblock \showarticletitle{Efficient Personalized Federated Learning via Sparse Model-Adaptation}. In \bibinfo{booktitle}{\emph{Proc. {ICML}}}, Vol.~\bibinfo{volume}{202}. \bibinfo{publisher}{{PMLR}}, \bibinfo{address}{Honolulu, Hawaii, {USA}}, \bibinfo{pages}{5234--5256}.
\newblock


\bibitem[Chen et~al\mbox{.}(2021)]%
        {FedMatch}
\bibfield{author}{\bibinfo{person}{Jiangui Chen} {et~al\mbox{.}}} \bibinfo{year}{2021}\natexlab{}.
\newblock \showarticletitle{FedMatch: Federated Learning Over Heterogeneous Question Answering Data}. In \bibinfo{booktitle}{\emph{Proc. {CIKM}}}. \bibinfo{publisher}{{ACM}}, \bibinfo{address}{virtual}, \bibinfo{pages}{181--190}.
\newblock


\bibitem[Cheng et~al\mbox{.}(2021)]%
        {FedGEMS}
\bibfield{author}{\bibinfo{person}{Sijie Cheng} {et~al\mbox{.}}} \bibinfo{year}{2021}\natexlab{}.
\newblock \showarticletitle{FedGEMS: Federated Learning of Larger Server Models via Selective Knowledge Fusion}.
\newblock \bibinfo{journal}{\emph{CoRR}}  \bibinfo{volume}{abs/2110.11027} (\bibinfo{year}{2021}).
\newblock


\bibitem[Cho et~al\mbox{.}(2022)]%
        {Fed-ET}
\bibfield{author}{\bibinfo{person}{Yae~Jee Cho} {et~al\mbox{.}}} \bibinfo{year}{2022}\natexlab{}.
\newblock \showarticletitle{Heterogeneous Ensemble Knowledge Transfer for Training Large Models in Federated Learning}. In \bibinfo{booktitle}{\emph{Proc. {IJCAI}}}. \bibinfo{publisher}{ijcai.org}, \bibinfo{address}{virtual}, \bibinfo{pages}{2881--2887}.
\newblock


\bibitem[Collins et~al\mbox{.}(2021)]%
        {FedRep}
\bibfield{author}{\bibinfo{person}{Liam Collins} {et~al\mbox{.}}} \bibinfo{year}{2021}\natexlab{}.
\newblock \showarticletitle{Exploiting Shared Representations for Personalized Federated Learning}. In \bibinfo{booktitle}{\emph{Proc. {ICML}}}, Vol.~\bibinfo{volume}{139}. \bibinfo{publisher}{{PMLR}}, \bibinfo{address}{virtual}, \bibinfo{pages}{2089--2099}.
\newblock


\bibitem[Diao(2021)]%
        {HeteroFL}
\bibfield{author}{\bibinfo{person}{Enmao Diao}.} \bibinfo{year}{2021}\natexlab{}.
\newblock \showarticletitle{HeteroFL: Computation and Communication Efficient Federated Learning for Heterogeneous Clients}. In \bibinfo{booktitle}{\emph{Proc. ICLR}}. \bibinfo{publisher}{OpenReview.net}, \bibinfo{address}{Virtual Event, Austria}, \bibinfo{pages}{1}.
\newblock


\bibitem[Dun et~al\mbox{.}(2023)]%
        {FedJETs}
\bibfield{author}{\bibinfo{person}{Chen Dun} {et~al\mbox{.}}} \bibinfo{year}{2023}\natexlab{}.
\newblock \bibinfo{title}{FedJETs: Efficient Just-In-Time Personalization with Federated Mixture of Experts}.
\newblock
\newblock
\showeprint[arxiv]{2306.08586}~[cs.LG]


\bibitem[Guo et~al\mbox{.}(2021)]%
        {PFL-MoE}
\bibfield{author}{\bibinfo{person}{Binbin Guo} {et~al\mbox{.}}} \bibinfo{year}{2021}\natexlab{}.
\newblock \showarticletitle{PFL-MoE: Personalized Federated Learning Based on Mixture of Experts}. In \bibinfo{booktitle}{\emph{Proc. APWeb-WAIM}}, Vol.~\bibinfo{volume}{12858}. \bibinfo{publisher}{Springer}, \bibinfo{address}{Guangzhou, China}, \bibinfo{pages}{480--486}.
\newblock


\bibitem[He et~al\mbox{.}(2020)]%
        {FedGKT}
\bibfield{author}{\bibinfo{person}{Chaoyang He} {et~al\mbox{.}}} \bibinfo{year}{2020}\natexlab{}.
\newblock \showarticletitle{Group Knowledge Transfer: Federated Learning of Large CNNs at the Edge}. In \bibinfo{booktitle}{\emph{Proc. NeurIPS}}. \bibinfo{publisher}{{}}, \bibinfo{address}{virtual}.
\newblock


\bibitem[Horv{\'a}th(2021)]%
        {FjORD}
\bibfield{author}{\bibinfo{person}{S. Horv{\'a}th}.} \bibinfo{year}{2021}\natexlab{}.
\newblock \showarticletitle{Fj{ORD}: Fair and Accurate Federated Learning under heterogeneous targets with Ordered Dropout}. In \bibinfo{booktitle}{\emph{Proc. {NIPS}}}. \bibinfo{publisher}{OpenReview.net}, \bibinfo{address}{Virtual}, \bibinfo{pages}{12876--12889}.
\newblock


\bibitem[Huang et~al\mbox{.}(2022a)]%
        {FSFL}
\bibfield{author}{\bibinfo{person}{Wenke Huang} {et~al\mbox{.}}} \bibinfo{year}{2022}\natexlab{a}.
\newblock \showarticletitle{Few-Shot Model Agnostic Federated Learning}. In \bibinfo{booktitle}{\emph{Proc. {MM}}}. \bibinfo{publisher}{{ACM}}, \bibinfo{address}{Lisboa, Portugal}, \bibinfo{pages}{7309--7316}.
\newblock


\bibitem[Huang et~al\mbox{.}(2022b)]%
        {FCCL}
\bibfield{author}{\bibinfo{person}{Wenke Huang} {et~al\mbox{.}}} \bibinfo{year}{2022}\natexlab{b}.
\newblock \showarticletitle{Learn from Others and Be Yourself in Heterogeneous Federated Learning}. In \bibinfo{booktitle}{\emph{Proc. {CVPR}}}. \bibinfo{publisher}{{IEEE}}, \bibinfo{address}{virtual}, \bibinfo{pages}{10133--10143}.
\newblock


\bibitem[Itahara et~al\mbox{.}(2023)]%
        {DS-FL}
\bibfield{author}{\bibinfo{person}{Sohei Itahara} {et~al\mbox{.}}} \bibinfo{year}{2023}\natexlab{}.
\newblock \showarticletitle{Distillation-Based Semi-Supervised Federated Learning for Communication-Efficient Collaborative Training With Non-IID Private Data}.
\newblock \bibinfo{journal}{\emph{{IEEE} Trans. Mob. Comput.}} \bibinfo{volume}{22}, \bibinfo{number}{1} (\bibinfo{year}{2023}), \bibinfo{pages}{191--205}.
\newblock


\bibitem[Jang et~al\mbox{.}(2022)]%
        {FedClassAvg}
\bibfield{author}{\bibinfo{person}{Jaehee Jang} {et~al\mbox{.}}} \bibinfo{year}{2022}\natexlab{}.
\newblock \showarticletitle{FedClassAvg: Local Representation Learning for Personalized Federated Learning on Heterogeneous Neural Networks}. In \bibinfo{booktitle}{\emph{Proc. {ICPP}}}. \bibinfo{publisher}{{ACM}}, \bibinfo{address}{virtual}, \bibinfo{pages}{76:1--76:10}.
\newblock


\bibitem[Jeong et~al\mbox{.}(2018)]%
        {FD}
\bibfield{author}{\bibinfo{person}{Eunjeong Jeong} {et~al\mbox{.}}} \bibinfo{year}{2018}\natexlab{}.
\newblock \showarticletitle{Communication-Efficient On-Device Machine Learning: Federated Distillation and Augmentation under Non-IID Private Data}. In \bibinfo{booktitle}{\emph{Proc. NeurIPS Workshop on Machine Learning on the Phone and other Consumer Devices}}. \bibinfo{publisher}{{}}, \bibinfo{address}{virtual}.
\newblock


\bibitem[Kairouz et~al\mbox{.}(2021)]%
        {1w-survey}
\bibfield{author}{\bibinfo{person}{Peter Kairouz} {et~al\mbox{.}}} \bibinfo{year}{2021}\natexlab{}.
\newblock \showarticletitle{Advances and Open Problems in Federated Learning}.
\newblock \bibinfo{journal}{\emph{Foundations and Trends in Machine Learning}} \bibinfo{volume}{14}, \bibinfo{number}{1--2} (\bibinfo{year}{2021}), \bibinfo{pages}{1--210}.
\newblock


\bibitem[Krizhevsky et~al\mbox{.}(2009)]%
        {cifar}
\bibfield{author}{\bibinfo{person}{Alex Krizhevsky} {et~al\mbox{.}}} \bibinfo{year}{2009}\natexlab{}.
\newblock \bibinfo{booktitle}{\emph{Learning multiple layers of features from tiny images}}.
\newblock \bibinfo{publisher}{Toronto, ON, Canada}, \bibinfo{address}{{}}.
\newblock


\bibitem[Li and Wang(2019)]%
        {FedMD}
\bibfield{author}{\bibinfo{person}{Daliang Li} {and} \bibinfo{person}{Junpu Wang}.} \bibinfo{year}{2019}\natexlab{}.
\newblock \showarticletitle{FedMD: Heterogenous Federated Learning via Model Distillation}. In \bibinfo{booktitle}{\emph{Proc. NeurIPS Workshop}}. \bibinfo{publisher}{{}}, \bibinfo{address}{virtual}.
\newblock


\bibitem[Li et~al\mbox{.}(2021)]%
        {FedKT}
\bibfield{author}{\bibinfo{person}{Qinbin Li} {et~al\mbox{.}}} \bibinfo{year}{2021}\natexlab{}.
\newblock \showarticletitle{Practical One-Shot Federated Learning for Cross-Silo Setting}. In \bibinfo{booktitle}{\emph{Proc. {IJCAI}}}. \bibinfo{publisher}{ijcai.org}, \bibinfo{address}{virtual}, \bibinfo{pages}{1484--1490}.
\newblock


\bibitem[Li et~al\mbox{.}(2020)]%
        {xxl-moe}
\bibfield{author}{\bibinfo{person}{Xiaoxiao Li} {et~al\mbox{.}}} \bibinfo{year}{2020}\natexlab{}.
\newblock \showarticletitle{Multi-site fMRI analysis using privacy-preserving federated learning and domain adaptation: {ABIDE} results}.
\newblock \bibinfo{journal}{\emph{Medical Image Anal.}}  \bibinfo{volume}{65} (\bibinfo{year}{2020}), \bibinfo{pages}{101765}.
\newblock


\bibitem[Liang et~al\mbox{.}(2020)]%
        {LG-FedAvg}
\bibfield{author}{\bibinfo{person}{Paul~Pu Liang} {et~al\mbox{.}}} \bibinfo{year}{2020}\natexlab{}.
\newblock \showarticletitle{Think locally, act globally: Federated learning with local and global representations}.
\newblock \bibinfo{journal}{\emph{arXiv preprint arXiv:2001.01523}} \bibinfo{volume}{1}, \bibinfo{number}{1} (\bibinfo{year}{2020}).
\newblock


\bibitem[Lin et~al\mbox{.}(2020)]%
        {FedDF}
\bibfield{author}{\bibinfo{person}{Tao Lin} {et~al\mbox{.}}} \bibinfo{year}{2020}\natexlab{}.
\newblock \showarticletitle{Ensemble Distillation for Robust Model Fusion in Federated Learning}. In \bibinfo{booktitle}{\emph{Proc. NeurIPS}}. \bibinfo{publisher}{{}}, \bibinfo{address}{virtual}.
\newblock


\bibitem[Liu et~al\mbox{.}(2022)]%
        {CHFL}
\bibfield{author}{\bibinfo{person}{Chang Liu} {et~al\mbox{.}}} \bibinfo{year}{2022}\natexlab{}.
\newblock \showarticletitle{Completely Heterogeneous Federated Learning}.
\newblock \bibinfo{journal}{\emph{CoRR}}  \bibinfo{volume}{abs/2210.15865} (\bibinfo{year}{2022}).
\newblock


\bibitem[Lu et~al\mbox{.}(2022)]%
        {HFL}
\bibfield{author}{\bibinfo{person}{Xiaofeng Lu} {et~al\mbox{.}}} \bibinfo{year}{2022}\natexlab{}.
\newblock \showarticletitle{Heterogeneous Model Fusion Federated Learning Mechanism Based on Model Mapping}.
\newblock \bibinfo{journal}{\emph{{IEEE} Internet Things J.}} \bibinfo{volume}{9}, \bibinfo{number}{8} (\bibinfo{year}{2022}), \bibinfo{pages}{6058--6068}.
\newblock


\bibitem[Luo et~al\mbox{.}(2019)]%
        {switchnorm}
\bibfield{author}{\bibinfo{person}{Ping Luo} {et~al\mbox{.}}} \bibinfo{year}{2019}\natexlab{}.
\newblock \showarticletitle{Differentiable Learning-to-Normalize via Switchable Normalization}. In \bibinfo{booktitle}{\emph{Proc. {ICLR}}}. \bibinfo{publisher}{OpenReview.net}, \bibinfo{address}{New Orleans, LA, USA}, \bibinfo{pages}{1}.
\newblock


\bibitem[Makhija et~al\mbox{.}(2022)]%
        {FedHeNN}
\bibfield{author}{\bibinfo{person}{Disha Makhija} {et~al\mbox{.}}} \bibinfo{year}{2022}\natexlab{}.
\newblock \showarticletitle{Architecture Agnostic Federated Learning for Neural Networks}. In \bibinfo{booktitle}{\emph{Proc. {ICML}}}, Vol.~\bibinfo{volume}{162}. \bibinfo{publisher}{{PMLR}}, \bibinfo{address}{virtual}, \bibinfo{pages}{14860--14870}.
\newblock


\bibitem[McMahan et~al\mbox{.}(2017)]%
        {FedAvg}
\bibfield{author}{\bibinfo{person}{Brendan McMahan} {et~al\mbox{.}}} \bibinfo{year}{2017}\natexlab{}.
\newblock \showarticletitle{Communication-Efficient Learning of Deep Networks from Decentralized Data}. In \bibinfo{booktitle}{\emph{Proc. {AISTATS}}}, Vol.~\bibinfo{volume}{54}. \bibinfo{publisher}{{PMLR}}, \bibinfo{address}{Fort Lauderdale, FL, {USA}}, \bibinfo{pages}{1273--1282}.
\newblock


\bibitem[Nguyen et~al\mbox{.}(2023)]%
        {FedKEM}
\bibfield{author}{\bibinfo{person}{Duy~Phuong Nguyen} {et~al\mbox{.}}} \bibinfo{year}{2023}\natexlab{}.
\newblock \showarticletitle{Enhancing Heterogeneous Federated Learning with Knowledge Extraction and Multi-Model Fusion}. In \bibinfo{booktitle}{\emph{Proc. {SC} Workshop}}. \bibinfo{publisher}{{ACM}}, \bibinfo{address}{Denver, CO, USA}, \bibinfo{pages}{36--43}.
\newblock


\bibitem[Oh et~al\mbox{.}(2022)]%
        {FedBABU}
\bibfield{author}{\bibinfo{person}{Jaehoon Oh} {et~al\mbox{.}}} \bibinfo{year}{2022}\natexlab{}.
\newblock \showarticletitle{FedBABU: Toward Enhanced Representation for Federated Image Classification}. In \bibinfo{booktitle}{\emph{Proc. {ICLR}}}. \bibinfo{publisher}{OpenReview.net}, \bibinfo{address}{virtual}.
\newblock


\bibitem[Park et~al\mbox{.}(2023)]%
        {KRR-KD}
\bibfield{author}{\bibinfo{person}{Sejun Park} {et~al\mbox{.}}} \bibinfo{year}{2023}\natexlab{}.
\newblock \showarticletitle{Towards Understanding Ensemble Distillation in Federated Learning}. In \bibinfo{booktitle}{\emph{Proc. {ICML}}}, Vol.~\bibinfo{volume}{202}. \bibinfo{publisher}{{PMLR}}, \bibinfo{address}{Honolulu, Hawaii, {USA}}, \bibinfo{pages}{27132--27187}.
\newblock


\bibitem[Pillutla et~al\mbox{.}(2022)]%
        {FedAlt/FedSim}
\bibfield{author}{\bibinfo{person}{Krishna Pillutla} {et~al\mbox{.}}} \bibinfo{year}{2022}\natexlab{}.
\newblock \showarticletitle{Federated Learning with Partial Model Personalization}. In \bibinfo{booktitle}{\emph{Proc. {ICML}}}, Vol.~\bibinfo{volume}{162}. \bibinfo{publisher}{{PMLR}}, \bibinfo{address}{virtual}, \bibinfo{pages}{17716--17758}.
\newblock


\bibitem[Qin et~al\mbox{.}(2023)]%
        {FedAPEN}
\bibfield{author}{\bibinfo{person}{Zhen Qin} {et~al\mbox{.}}} \bibinfo{year}{2023}\natexlab{}.
\newblock \showarticletitle{FedAPEN: Personalized Cross-silo Federated Learning with Adaptability to Statistical Heterogeneity}. In \bibinfo{booktitle}{\emph{Proc. {KDD}}}. \bibinfo{publisher}{{ACM}}, \bibinfo{address}{Long Beach, CA, USA}, \bibinfo{pages}{1954--1964}.
\newblock


\bibitem[Reisser et~al\mbox{.}(2021)]%
        {FedMix}
\bibfield{author}{\bibinfo{person}{Matthias Reisser} {et~al\mbox{.}}} \bibinfo{year}{2021}\natexlab{}.
\newblock \showarticletitle{Federated Mixture of Experts}.
\newblock \bibinfo{journal}{\emph{CoRR}}  \bibinfo{volume}{abs/2107.06724} (\bibinfo{year}{2021}), \bibinfo{pages}{1}.
\newblock


\bibitem[Ruder(2016)]%
        {SGD}
\bibfield{author}{\bibinfo{person}{Sebastian Ruder}.} \bibinfo{year}{2016}\natexlab{}.
\newblock \showarticletitle{An overview of gradient descent optimization algorithms}.
\newblock \bibinfo{journal}{\emph{CoRR}}  \bibinfo{volume}{abs/1609.04747} (\bibinfo{year}{2016}), \bibinfo{pages}{1}.
\newblock


\bibitem[Sattler et~al\mbox{.}(2021)]%
        {FEDAUX}
\bibfield{author}{\bibinfo{person}{Felix Sattler} {et~al\mbox{.}}} \bibinfo{year}{2021}\natexlab{}.
\newblock \showarticletitle{FEDAUX: Leveraging Unlabeled Auxiliary Data in Federated Learning}.
\newblock \bibinfo{journal}{\emph{{IEEE} Trans. Neural Networks Learn. Syst.}} \bibinfo{volume}{1}, \bibinfo{number}{1} (\bibinfo{year}{2021}), \bibinfo{pages}{1--13}.
\newblock


\bibitem[Sattler et~al\mbox{.}(2022)]%
        {CFD}
\bibfield{author}{\bibinfo{person}{Felix Sattler} {et~al\mbox{.}}} \bibinfo{year}{2022}\natexlab{}.
\newblock \showarticletitle{{CFD:} Communication-Efficient Federated Distillation via Soft-Label Quantization and Delta Coding}.
\newblock \bibinfo{journal}{\emph{{IEEE} Trans. Netw. Sci. Eng.}} \bibinfo{volume}{9}, \bibinfo{number}{4} (\bibinfo{year}{2022}), \bibinfo{pages}{2025--2038}.
\newblock


\bibitem[Shamsian et~al\mbox{.}(2021)]%
        {pFedHN}
\bibfield{author}{\bibinfo{person}{Aviv Shamsian} {et~al\mbox{.}}} \bibinfo{year}{2021}\natexlab{}.
\newblock \showarticletitle{Personalized Federated Learning using Hypernetworks}. In \bibinfo{booktitle}{\emph{Proc. {ICML}}}, Vol.~\bibinfo{volume}{139}. \bibinfo{publisher}{{PMLR}}, \bibinfo{address}{virtual}, \bibinfo{pages}{9489--9502}.
\newblock


\bibitem[Shen et~al\mbox{.}(2020)]%
        {FML}
\bibfield{author}{\bibinfo{person}{Tao Shen} {et~al\mbox{.}}} \bibinfo{year}{2020}\natexlab{}.
\newblock \showarticletitle{Federated Mutual Learning}.
\newblock \bibinfo{journal}{\emph{CoRR}}  \bibinfo{volume}{abs/2006.16765} (\bibinfo{year}{2020}).
\newblock


\bibitem[Tan et~al\mbox{.}(2022)]%
        {FedProto}
\bibfield{author}{\bibinfo{person}{Yue Tan} {et~al\mbox{.}}} \bibinfo{year}{2022}\natexlab{}.
\newblock \showarticletitle{FedProto: Federated Prototype Learning across Heterogeneous Clients}. In \bibinfo{booktitle}{\emph{Proc. {AAAI}}}. \bibinfo{publisher}{{AAAI} Press}, \bibinfo{address}{virtual}, \bibinfo{pages}{8432--8440}.
\newblock


\bibitem[van~der Maaten and Hinton(2008)]%
        {TSNE-JMLR}
\bibfield{author}{\bibinfo{person}{Laurens van~der Maaten} {and} \bibinfo{person}{Geoffrey Hinton}.} \bibinfo{year}{2008}\natexlab{}.
\newblock \showarticletitle{Visualizing Data using t-SNE}.
\newblock \bibinfo{journal}{\emph{Journal of Machine Learning Research}} \bibinfo{volume}{9}, \bibinfo{number}{86} (\bibinfo{year}{2008}), \bibinfo{pages}{2579--2605}.
\newblock


\bibitem[Wang et~al\mbox{.}(2023)]%
        {pFedHR}
\bibfield{author}{\bibinfo{person}{Jiaqi Wang} {et~al\mbox{.}}} \bibinfo{year}{2023}\natexlab{}.
\newblock \showarticletitle{Towards Personalized Federated Learning via Heterogeneous Model Reassembly}. In \bibinfo{booktitle}{\emph{Proc. {NeurIPS}}}. \bibinfo{publisher}{OpenReview.net}, \bibinfo{address}{New Orleans, Louisiana, USA}, \bibinfo{pages}{13}.
\newblock


\bibitem[Wu et~al\mbox{.}(2022)]%
        {FedKD}
\bibfield{author}{\bibinfo{person}{Chuhan Wu} {et~al\mbox{.}}} \bibinfo{year}{2022}\natexlab{}.
\newblock \showarticletitle{Communication-efficient federated learning via knowledge distillation}.
\newblock \bibinfo{journal}{\emph{Nature Communications}} \bibinfo{volume}{13}, \bibinfo{number}{1} (\bibinfo{year}{2022}), \bibinfo{pages}{2032}.
\newblock


\bibitem[Yi et~al\mbox{.}(2023)]%
        {FedGH}
\bibfield{author}{\bibinfo{person}{Liping Yi}, \bibinfo{person}{Gang Wang}, \bibinfo{person}{Xiaoguang Liu}, \bibinfo{person}{Zhuan Shi}, {and} \bibinfo{person}{Han Yu}.} \bibinfo{year}{2023}\natexlab{}.
\newblock \showarticletitle{FedGH: Heterogeneous Federated Learning with Generalized Global Header}. In \bibinfo{booktitle}{\emph{Proceedings of the 31st ACM International Conference on Multimedia (ACM MM'23)}}. \bibinfo{publisher}{ACM}, \bibinfo{address}{Canada}, \bibinfo{pages}{11}.
\newblock


\bibitem[Yu et~al\mbox{.}(2021)]%
        {Fed2}
\bibfield{author}{\bibinfo{person}{Fuxun Yu} {et~al\mbox{.}}} \bibinfo{year}{2021}\natexlab{}.
\newblock \showarticletitle{Fed2: Feature-Aligned Federated Learning}. In \bibinfo{booktitle}{\emph{Proc. {KDD}}}. \bibinfo{publisher}{{ACM}}, \bibinfo{address}{virtual}, \bibinfo{pages}{2066--2074}.
\newblock


\bibitem[Yu et~al\mbox{.}(2022)]%
        {FedKEMF}
\bibfield{author}{\bibinfo{person}{Sixing Yu} {et~al\mbox{.}}} \bibinfo{year}{2022}\natexlab{}.
\newblock \showarticletitle{Resource-aware Federated Learning using Knowledge Extraction and Multi-model Fusion}.
\newblock \bibinfo{journal}{\emph{CoRR}}  \bibinfo{volume}{abs/2208.07978} (\bibinfo{year}{2022}).
\newblock


\bibitem[Zec et~al\mbox{.}(2020)]%
        {FL-MoE2}
\bibfield{author}{\bibinfo{person}{Edvin~Listo Zec} {et~al\mbox{.}}} \bibinfo{year}{2020}\natexlab{}.
\newblock \showarticletitle{Federated learning using a mixture of experts}.
\newblock \bibinfo{journal}{\emph{CoRR}}  \bibinfo{volume}{abs/2010.02056} (\bibinfo{year}{2020}), \bibinfo{pages}{1}.
\newblock


\bibitem[Zhang et~al\mbox{.}(2021)]%
        {KT-pFL}
\bibfield{author}{\bibinfo{person}{Jie Zhang} {et~al\mbox{.}}} \bibinfo{year}{2021}\natexlab{}.
\newblock \showarticletitle{Parameterized Knowledge Transfer for Personalized Federated Learning}. In \bibinfo{booktitle}{\emph{Proc. NeurIPS}}. \bibinfo{publisher}{OpenReview.net}, \bibinfo{address}{virtual}, \bibinfo{pages}{10092--10104}.
\newblock


\bibitem[Zhang et~al\mbox{.}(2023a)]%
        {FedCP}
\bibfield{author}{\bibinfo{person}{Jianqing Zhang} {et~al\mbox{.}}} \bibinfo{year}{2023}\natexlab{a}.
\newblock \showarticletitle{FedCP: Separating Feature Information for Personalized Federated Learning via Conditional Policy}. In \bibinfo{booktitle}{\emph{Proc. {KDD}}}. \bibinfo{publisher}{{ACM}}, \bibinfo{address}{Long Beach, CA, USA}, \bibinfo{pages}{1}.
\newblock


\bibitem[Zhang et~al\mbox{.}(2023b)]%
        {FedGD}
\bibfield{author}{\bibinfo{person}{Jie Zhang} {et~al\mbox{.}}} \bibinfo{year}{2023}\natexlab{b}.
\newblock \showarticletitle{Towards Data-Independent Knowledge Transfer in Model-Heterogeneous Federated Learning}.
\newblock \bibinfo{journal}{\emph{{IEEE} Trans. Computers}} \bibinfo{volume}{72}, \bibinfo{number}{10} (\bibinfo{year}{2023}), \bibinfo{pages}{2888--2901}.
\newblock


\bibitem[Zhang et~al\mbox{.}(2022)]%
        {FedZKT}
\bibfield{author}{\bibinfo{person}{Lan Zhang} {et~al\mbox{.}}} \bibinfo{year}{2022}\natexlab{}.
\newblock \showarticletitle{FedZKT: Zero-Shot Knowledge Transfer towards Resource-Constrained Federated Learning with Heterogeneous On-Device Models}. In \bibinfo{booktitle}{\emph{Proc. {ICDCS}}}. \bibinfo{publisher}{{IEEE}}, \bibinfo{address}{virtual}, \bibinfo{pages}{928--938}.
\newblock


\bibitem[Zhang and Sabuncu(2018)]%
        {CEloss}
\bibfield{author}{\bibinfo{person}{Zhilu Zhang} {and} \bibinfo{person}{Mert~R. Sabuncu}.} \bibinfo{year}{2018}\natexlab{}.
\newblock \showarticletitle{Generalized Cross Entropy Loss for Training Deep Neural Networks with Noisy Labels}. In \bibinfo{booktitle}{\emph{Proc. {NeurIPS}}}. \bibinfo{publisher}{Curran Associates Inc.}, \bibinfo{address}{Montr{\'{e}}al, Canada}, \bibinfo{pages}{8792--8802}.
\newblock


\bibitem[Zhu et~al\mbox{.}(2021)]%
        {FedGen}
\bibfield{author}{\bibinfo{person}{Zhuangdi Zhu} {et~al\mbox{.}}} \bibinfo{year}{2021}\natexlab{}.
\newblock \showarticletitle{Data-Free Knowledge Distillation for Heterogeneous Federated Learning}. In \bibinfo{booktitle}{\emph{Proc. {ICML}}}, Vol.~\bibinfo{volume}{139}. \bibinfo{publisher}{{PMLR}}, \bibinfo{address}{virtual}, \bibinfo{pages}{12878--12889}.
\newblock


\bibitem[Zhu et~al\mbox{.}(2022)]%
        {FedResCuE}
\bibfield{author}{\bibinfo{person}{Zhuangdi Zhu} {et~al\mbox{.}}} \bibinfo{year}{2022}\natexlab{}.
\newblock \showarticletitle{Resilient and Communication Efficient Learning for Heterogeneous Federated Systems}. In \bibinfo{booktitle}{\emph{Proc. {ICML}}}, Vol.~\bibinfo{volume}{162}. \bibinfo{publisher}{{PMLR}}, \bibinfo{address}{virtual}, \bibinfo{pages}{27504--27526}.
\newblock


\end{thebibliography}

\appendix
\onecolumn

\section{Pseudo codes of \methodname{}}\label{app:pseudo-codes}
\begin{algorithm}[h]
  \caption{\methodname{}}
  \label{alg:FedMoE}
  \textbf{Input}: $N$, total number of clients; $K$, number of sampled clients in one round; $T$, number of rounds; $\eta_\theta$, learning rate of homogeneous feature extractor; $\eta_\omega$, learning rate of local heterogeneous models; $\eta_\varphi$, learning rate of local homogeneous gating network. \\
     Randomly initialize the global homogeneous feature extractor $\mathcal{G}(\theta^\mathbf{0})$, local personalized heterogeneous models $[\mathcal{F}_0(\omega_0^0),\mathcal{F}_1(\omega_1^0),\ldots,\mathcal{F}_k(\omega_k^0),\ldots,\mathcal{F}_{N-1}(\omega_{N-1}^0)]$ and local homogeneous gating networks $\mathcal{H}(\varphi^\mathbf{0})$. \\
    \For{$t=1$ {\bfseries to} $T-1$}{
     // \textbf{Server Side}: \\
         $\boldsymbol{\mathcal{S}}^t \gets$  Randomly sample $K\leqslant N$ clients to join FL; \\
         Broadcast the global homogeneous feature extractor $\theta^{t-1}$ to sampled $K$ clients; \\ 
         $\theta_k^t \gets$ \textbf{Client Update}$(\theta^{t-1})$; \\
         \begin{tcolorbox}[colback=ylp_color2,
                  colframe=ylp_color1,
                  width=7.25cm,
                  height=1cm,
                  arc=1mm, auto outer arc,
                  boxrule=1pt,
                  left=0pt,right=0pt,top=0pt,bottom=0pt,
                 ]
         \textbf{/* Aggregate Homogeneous Feature Extractors */} \\
        $\theta^t=\sum_{k \in \boldsymbol{\mathcal{S}}^t} \frac{n_k}{n} \theta_k^t$. \\
        \end{tcolorbox}
         // \textbf{Client Update}:  \\
         Receive the global homogeneous feature extractor $\theta^{t-1}$ from the server; \\
            \For{$k \in \boldsymbol{\mathcal{S}}^t$}{
                  \For{$(\boldsymbol{x}_i,y_i)\in D_k$}{
         \begin{tcolorbox}[colback=ylp_color2,
                  colframe=ylp_color1,
                  width=6.25cm,
                  height=4.45cm,
                  arc=1mm, auto outer arc,
                  boxrule=1pt,
                  left=0pt,right=0pt,top=0pt,bottom=0pt,
                 ]
                 \textbf{/* Local MoE Training */} \\
                    $\boldsymbol{\mathcal{R}}_{k, i}^{\mathcal{G}, t}=\mathcal{G}(\boldsymbol{x}_i ; \theta^{t-1})$;
                    $\boldsymbol{\mathcal{R}}_{k, i}^{\mathcal{F}_{k},t}=\mathcal{F}_k^{e x}(\boldsymbol{x}_i ; \omega_k^{e x, t-1})$; \\
                    $[\alpha_{k, i}^{\mathcal{G}, t}, \alpha_{k, i}^{\mathcal{F}_{k}, t}]=\mathcal{H}(\boldsymbol{x}_i ; \varphi_k^{t-1})$; \\
                    $\boldsymbol{\mathcal{R}}_{k, i}^t=\alpha_{k, i}^{\mathcal{G}, t} \cdot \boldsymbol{\mathcal{R}}_{k, i}^{\mathcal{G}, t}+\alpha_{k, i}^{\mathcal{F}_{k}, t} \cdot \boldsymbol{\mathcal{R}}_{k, i}^{\mathcal{F}_{k},t}$; \\
                    $\hat{y}_i=\mathcal{F}_k^{h d}(\boldsymbol{\mathcal{R}}_{k, i}^t ; \omega_k^{h d, t-1})$; \\
                    $\ell_i=C E(\hat{y}_i, y_i)$; \\
                    $\theta_k^t \gets \theta^{t-1}-\eta_\theta \nabla \ell_i$; \\
                    $\omega_k^t \gets \omega_k^{t-1}-\eta_\omega \nabla \ell_i$; \\
                    $\varphi_k^t \gets \varphi_k^{t-1}-\eta_{\varphi} \nabla \ell_i$; 
                     \end{tcolorbox}
                  }
                Upload trained local homogeneous feature extractor $\theta_k^t$ to the server.
            }
    }
     \textbf{Return} personalized heterogeneous local complete models 
      \begin{tcolorbox}[colback=ylp_color2,
                  colframe=ylp_color1,
                  width=8.4cm,
                  height=0.75cm,
                  arc=1mm, auto outer arc,
                  boxrule=1pt,
                  left=0pt,right=0pt,top=0pt,bottom=0pt,
                 ]
     $[\{MoE(\mathcal{G}(\theta^{T-1}),\mathcal{F}_0^{ex}(\omega_0^{ex,T-1});\mathcal{H}(\varphi_0^{T-1})), \mathcal{F}_0^{hd}(\omega_0^{hd,T-1})\},\ldots]$.
      \end{tcolorbox}
\end{algorithm}

\section{Theoretical Proofs}\label{app:proof}
\subsection{Proof for Lemma~\ref{lemma:localtraining}}
An arbitrary client $k$'s local model $W$ can be updated by $W_{t+1}=W_t-\eta g_{W,t}$ in the (t+1)-th round, and following Assumption~\ref{assump:Lipschitz}, we can obtain
\begin{equation}
\begin{aligned}
 \mathcal{L}_{t E+1} &\leq \mathcal{L}_{t E+0}+\langle\nabla \mathcal{L}_{t E+0},(W_{t E+1}-W_{t E+0})\rangle+\frac{L_1}{2}\|W_{t E+1}-W_{t E+0}\|_2^2 \\
& =\mathcal{L}_{t E+0}-\eta\langle\nabla \mathcal{L}_{t E+0}, g_{W, t E+0}\rangle+\frac{L_1 \eta^2}{2}\|g_{W, t E+0}\|_2^2 .   
\end{aligned}
\end{equation}

Taking the expectation of both sides of the inequality concerning the random variable $\xi_{tE+0}$, we obtain
\begin{equation}
\begin{aligned}
 \mathbb{E}[\mathcal{L}_{t E+1}] &\leq \mathcal{L}_{t E+0}-\eta \mathbb{E}[\langle\nabla \mathcal{L}_{t E+0}, g_{W, t E+0}\rangle]+\frac{L_1 \eta^2}{2} \mathbb{E}[\|g_{W, t E+0}\|_2^2] \\
& \stackrel{(a)}{=} \mathcal{L}_{t E+0}-\eta\|\nabla \mathcal{L}_{t E+0}\|_2^2+\frac{L_1 \eta^2}{2} \mathbb{E}[\|g_{W, t E+0}\|_2^2] \\
& \stackrel{(b)}{\leq} \mathcal{L}_{t E+0}-\eta\|\nabla \mathcal{L}_{t E+0}\|_2^2+\frac{L_1 \eta^2}{2}(\mathbb{E}[\|g_{W, t E+0}\|]_2^2+\operatorname{Var}(g_{W, t E+0})) \\
& \stackrel{(c)}{=} \mathcal{L}_{t E+0}-\eta\|\nabla \mathcal{L}_{t E+0}\|_2^2+\frac{L_1 \eta^2}{2}(\|\nabla \mathcal{L}_{t E+0}\|_2^2+\operatorname{Var}(g_{W, t E+0})) \\
& \stackrel{(d)}{\leq} \mathcal{L}_{t E+0}-\eta\|\nabla \mathcal{L}_{t E+0}\|_2^2+\frac{L_1 \eta^2}{2}(\|\nabla \mathcal{L}_{t E+0}\|_2^2+\sigma^2) \\
& =\mathcal{L}_{t E+0}+(\frac{L_1 \eta^2}{2}-\eta)\|\nabla \mathcal{L}_{t E+0}\|_2^2+\frac{L_1 \eta^2 \sigma^2}{2}.
\end{aligned}
\end{equation}

(a), (c), (d) follow \assum \ref{assump:Unbiased}. (b) follows $Var(x)=\mathbb{E}[x^2]-(\mathbb{E}[x]^2)$.

Taking the expectation of both sides of the inequality for the model $W$ over $E$ iterations, we obtain

\begin{equation}
\mathbb{E}[\mathcal{L}_{t E+1}] \leq \mathcal{L}_{t E+0}+(\frac{L_1 \eta^2}{2}-\eta) \sum_{e=1}^E\|\nabla \mathcal{L}_{t E+e}\|_2^2+\frac{L_1 E \eta^2 \sigma^2}{2} . 
\end{equation}

\subsection{Proof for Lemma~\ref{lemma:aggregation}}
\begin{equation}
\begin{aligned}
\mathcal{L}_{(t+1) E+0}& =\mathcal{L}_{(t+1) E}+\mathcal{L}_{(t+1) E+0}-\mathcal{L}_{(t+1) E} \\
& \stackrel{(a)}{\approx} \mathcal{L}_{(t+1) E}+\eta\|\theta_{(t+1) E+0}-\theta_{(t+1) E}\|_2^2 \\
& \stackrel{(b)}{\leq} \mathcal{L}_{(t+1) E}+\eta \delta^2.
\end{aligned}
\end{equation}

(a): we can use the gradient of parameter variations to approximate the loss variations, \emph{i.e.}, $\Delta\mathcal{L}\approx \eta\cdot \|\Delta \theta\|_2^2$. (b) follows \assum \ref{assump:BoundedVariation}.

Taking the expectation of both sides of the inequality to the random variable $\xi$, we obtain
\begin{equation}
    \mathbb{E}\left[\mathcal{L}_{(t+1)E+0}\right]\le\mathbb{E}\left[\mathcal{L}_{tE+1}\right]+{\eta\delta}^2.
\end{equation}

\subsection{Proof for Theorem~\ref{theorem:one-round}}
Substituting Lemma~\ref{lemma:localtraining} into the right side of Lemma~\ref{lemma:aggregation}'s inequality, we obtain
\begin{equation}\label{eq:theorem1}
\mathbb{E}[\mathcal{L}_{(t+1) E+0}] \leq \mathcal{L}_{t E+0}+(\frac{L_1 \eta^2}{2}-\eta) \sum_{e=0}^E\|\nabla \mathcal{L}_{t E+e}\|_2^2+\frac{L_1 E \eta^2 \sigma^2}{2}+\eta \delta^2.
\end{equation}

\subsection{Proof for Theorem~\ref{theorem:non-convex}}
Interchanging the left and right sides of Eq.~(\ref{eq:theorem1}), we obtain
\begin{equation}
\sum_{e=0}^E\|\nabla \mathcal{L}_{t E+e}\|_2^2 \leq \frac{\mathcal{L}_{t E+0}-\mathbb{E}[\mathcal{L}_{(t+1) E+0}]+\frac{L_1 E \eta^2 \sigma^2}{2}+\eta \delta^2}{\eta-\frac{L_1 \eta^2}{2}}.
\end{equation}

Taking the expectation of both sides of the inequality over rounds $t= [0, T-1]$ to $W$, we obtain
\begin{equation}
\frac{1}{T} \sum_{t=0}^{T-1} \sum_{e=0}^{E-1}\|\nabla \mathcal{L}_{t E+e}\|_2^2 \leq \frac{\frac{1}{T} \sum_{t=0}^{T-1}[\mathcal{L}_{t E+0}-\mathbb{E}[\mathcal{L}_{(t+1) E+0}]]+\frac{L_1 E \eta^2 \sigma^2}{2}+\eta \delta^2}{\eta-\frac{L_1 \eta^2}{2}}.
\end{equation}

Let $\Delta=\mathcal{L}_{t=0} - \mathcal{L}^* > 0$, then $\sum_{t=0}^{T-1}[\mathcal{L}_{t E+0}-\mathbb{E}[\mathcal{L}_{(t+1) E+0}]] \leq \Delta$, we can get 
\begin{equation}\label{eq:theorem2}
\frac{1}{T} \sum_{t=0}^{T-1} \sum_{e=0}^{E-1}\|\nabla \mathcal{L}_{t E+e}\|_2^2 \leq \frac{\frac{\Delta}{T}+\frac{L_1 E \eta^2 \sigma^2}{2}+\eta \delta^2}{\eta-\frac{L_1 \eta^2}{2}}.
\end{equation}

If the above equation converges to a constant $\epsilon$, \emph{i.e.},

\begin{equation}
\frac{\frac{\Delta}{T}+\frac{L_1 E \eta^2 \sigma^2}{2}+\eta \delta^2}{\eta-\frac{L_1 \eta^2}{2}}<\epsilon,
\end{equation}
then 
\begin{equation}
T>\frac{\Delta}{\epsilon(\eta-\frac{L_1 \eta^2}{2})-\frac{L_1 E \eta^2 \sigma^2}{2}-\eta \delta^2}.
\end{equation}

Since $T>0, \Delta>0$, we can get
\begin{equation}
\epsilon(\eta-\frac{L_1 \eta^2}{2})-\frac{L_1 E \eta^2 \sigma^2}{2}-\eta \delta^2>0.
\end{equation}

Solving the above inequality yields

\begin{equation}
\eta<\frac{2(\epsilon-\delta^2)}{L_1(\epsilon+E \sigma^2)}.
\end{equation}

Since $\epsilon,\ L_1,\ \sigma^2,\ \delta^2$ are all constants greater than 0, $\eta$ has solutions.
Therefore, when the learning rate $\eta$ satisfies the above condition, any client's local complete \hetero model can converge. Notice that the learning rate of the local complete \hetero model involves $\{\eta_\theta,\eta_\omega,\eta_\varphi\}$, so it's crucial to set reasonable them to ensure model convergence. Since all terms on the right side of Eq.~(\ref{eq:theorem2}) except for $\Delta/T$ are constants, $\Delta$ is also a constant, \methodname{}'s non-convex convergence rate is $\epsilon \sim \mathcal{O}(\frac{1}{T})$.

\section{More Experimental Details and Results}\label{app:experiment}

\begin{table}[h]
\centering
\caption{Structures of $5$ heterogeneous CNN models with $5 \times 5$ kernel size and $16$ or $32$ filters in convolutional layers.}
\resizebox{0.5\linewidth}{!}{%
\begin{tabular}{|l|c|c|c|c|c|}
\hline
Layer Name         & CNN-1    & CNN-2   & CNN-3   & CNN-4   & CNN-5   \\ \hline
Conv1              & 5$\times$5, 16   & 5$\times$5, 16  & 5$\times$5, 16  & 5$\times$5, 16  & 5$\times$5, 16  \\
Maxpool1              & 2$\times$2   & 2$\times$2  & 2$\times$2  & 2$\times$2  & 2$\times$2  \\
Conv2              & 5$\times$5, 32   & 5$\times$5, 16  & 5$\times$5, 32  & 5$\times$5, 32  & 5$\times$5, 32  \\
Maxpool2              & 2$\times$2   & 2$\times$2  & 2$\times$2  & 2$\times$2  & 2$\times$2  \\
FC1                & 2000     & 2000    & 1000    & 800     & 500     \\
FC2                & 500      & 500     & 500     & 500     & 500     \\
FC3                & 10/100   & 10/100  & 10/100  & 10/100  & 10/100  \\ \hline
model size & 10.00 MB & 6.92 MB & 5.04 MB & 3.81 MB & 2.55 MB \\ \hline
\end{tabular}%
}
\label{tab:model-structures}
\end{table}

\begin{figure*}[h]
\centering
\begin{minipage}[t]{0.3\linewidth}
\centering
\includegraphics[width=2in]{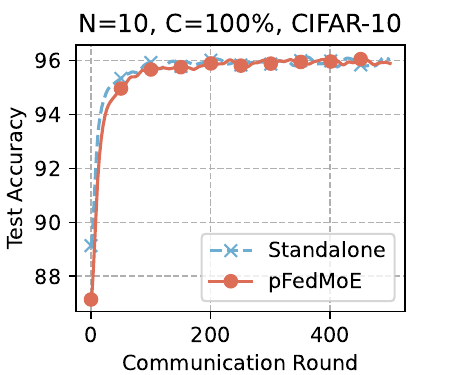}
\end{minipage}%
\begin{minipage}[t]{0.3\linewidth}
\centering
\includegraphics[width=2in]{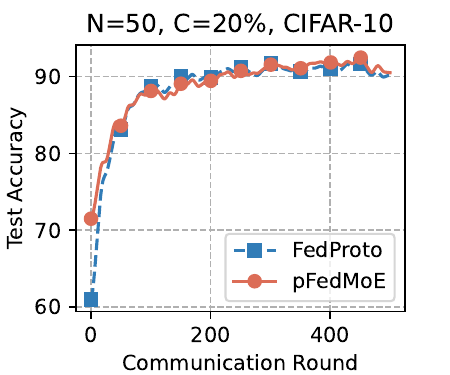}
\end{minipage}%
\begin{minipage}[t]{0.3\linewidth}
\centering
\includegraphics[width=2in]{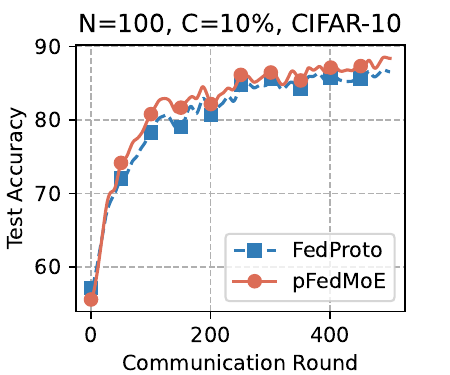}
\end{minipage}%

\begin{minipage}[t]{0.3\linewidth}
\centering
\includegraphics[width=2in]{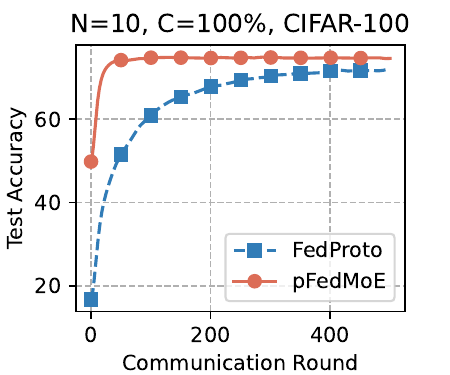}
\end{minipage}%
\begin{minipage}[t]{0.3\linewidth}
\centering
\includegraphics[width=2in]{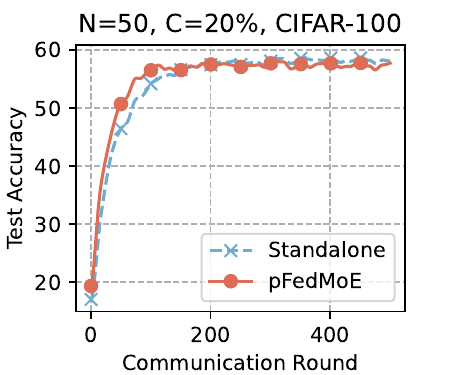}
\end{minipage}%
\begin{minipage}[t]{0.3\linewidth}
\centering
\includegraphics[width=2in]{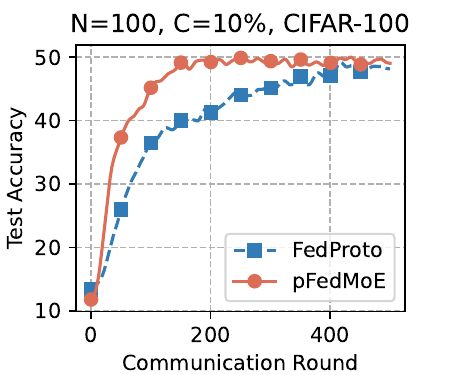}
\end{minipage}%
\caption{Average accuracy vs. communication rounds.}
\label{fig:compare-hetero-converge}
\end{figure*}

\end{document}